\documentclass[11pt]{article}
\usepackage{booktabs}
\usepackage[final]{acl}

\usepackage{times}
\usepackage{latexsym}

\usepackage[T1]{fontenc}

\usepackage[utf8]{inputenc}
\usepackage{amsmath}

\usepackage{microtype}

\usepackage{inconsolata}

\usepackage{graphicx}
\usepackage{subcaption}
\usepackage{amsfonts}
\title{SCOPE-RL: Optimizing Reasoning Paths Before and After Success}

\author{
\textbf{Xiaojian Liu}$^{1,2*}$, \textbf{Han Xu}$^{1*\dagger}$, \textbf{Jianqiang Xia}$^1$, \textbf{Zhixuan Li}$^1$, \textbf{Ke Xu}$^1$, \\
\textbf{Yiwei Dai}$^1$, \textbf{Xinran Chen}$^1$, \textbf{Changwo Wu}$^1$, \textbf{Yuchen Li}$^1$ \\
$^1$ Baidu Inc. \quad $^2$ Shandong University \\
\texttt{liux83596@gmail.com}, \texttt{xhbj66@gmail.com}
}

\usepackage{booktabs}
\usepackage{multirow}
\usepackage{graphicx} 
\usepackage{algorithm}
\usepackage{algorithmic}
\usepackage{tcolorbox}

\newcommand\blfootnote[1]{%
  \begingroup
  \renewcommand\thefootnote{}\footnote{#1}%
  \addtocounter{footnote}{-1}%
  \endgroup
}

\begin{document}
\maketitle

\blfootnote{$^*$Equal contribution. $^\dagger$Corresponding author.}

\begin{abstract}
Reinforcement learning with verifiable rewards (RLVR) optimizes LLMs using sparse verifiable final-answer rewards. This \emph{sparse anchor} reliably verifies whether a trajectory succeeds but provides no direct feedback on the reasoning path that produced it. \textbf{Before success}, prerequisite progress on hard problems receives no reward signal; \textbf{after success}, outcome rewards cannot distinguish well-organized correct trajectories from redundant or locally flawed ones. We introduce \textbf{SCOPE-RL} (Scaffolded Chain Optimization with Process Efficiency), a two-stage framework that densifies this anchor while retaining the GRPO update: \textbf{Adaptive Scaffolded RL} adds prefix-decomposed verifiable rewards on answer-hidden sub-question chains before success, and \textbf{Quality-Aware Process RL} applies correctness-gated process-shape rewards to refine correct trajectories after success. An expert-validated Step-Quality Evaluation Protocol evaluates useful-step density, error localization, and token efficiency beyond final-answer accuracy. On Qwen3-8B-Instruct trained on DAPO-Math and Big-Math, SCOPE-RL improves average accuracy by up to 11.2~pp and reduces reasoning tokens by up to 27.1\% over outcome-only GRPO; the gains hold under GSPO and on Qwen3-0.6B-Instruct, indicating that reward-signal densification is complementary to policy-update-level RLVR advances.Code and data are available at \url{https://github.com/tokencraft-lab/SCOPE-RL}.

\end{abstract}

\section{Introduction}
Reinforcement learning with verifiable rewards (RLVR) optimizes LLMs using sparse terminal rewards, typically final-answer correctness \citep{deepseek-r1, grpo, dapo, gspo, drgrpo}. This terminal anchor reliably certifies success but provides no direct feedback on the reasoning path, leading to two failure modes usually treated separately: training stalls on hard problems where outcome-only rollouts rarely succeed \citep{reasoningcurriculum, adarft, relift}, and models produce correct but inefficient reasoning on problems they can already solve \citep{whenmorethinkinghurts, ccc, smartthinker}. \textbf{Before success}, prerequisite progress is invisible unless it reaches the final answer, so near-misses provide little signal; \textbf{after success}, endpoint rewards cannot distinguish well-organized correct trajectories from redundant or locally flawed ones. Two diagnostic probes provide evidence for this interpretation (Figure~\ref{fig:motivation}; details in Appendix~\ref{app:motivation_probe}): scaffolded prompts reveal prerequisite signal that endpoint rewards leave unexposed, and among final-answer-correct rollouts outcome-only GRPO increases response length while retaining low-value content.

This sparse-anchor view reframes prior work as addressing different consequences of the same anchor: exploration-oriented methods reshape inputs, rollouts, or updates on hard problems \citep{e2h, reasoningcurriculum, rorl, adarft, cogdrift, ladder, pope, sage, stephint, evocot, luffy}, efficiency methods reduce overthinking \citep{thinkprune, smartthinker, ccc}, and process-supervision methods train PRMs, derive implicit rewards, or reassign dense feedback \citep{lightman2023, mathshepherd, cui2025process, progrs, prl, pure}. Less explored is how the rule-verifiable signal itself can be densified through independently verified sub-answer targets before success and correctness-gated process composition after success.

We instantiate this view as \textbf{SCOPE-RL} (Scaffolded Chain Optimization with Process Efficiency), a two-stage RLVR framework that densifies the sparse anchor while retaining the GRPO update. \textbf{Adaptive Scaffolded RL} (ASR) targets the before-success phase: for problems where outcome-only rollouts yield low success rates, it converts answer-hidden scaffolded sub-question chains into prefix-decomposed verifiable rewards, crediting partial progress without leaking answers. \textbf{Quality-Aware Process RL} (QPR) targets the after-success phase: on trajectories that reach the correct answer, it applies correctness-gated process-shape rewards that prefer useful, well-organized reasoning over redundant or locally flawed reasoning. The stages are naturally sequential---ASR first raises the supply of verified trajectories, after which QPR refines their process quality---and by design neither allows answer leakage nor brevity to substitute for verified correctness. To evaluate reasoning processes beyond final-answer accuracy, we introduce a \textbf{Step-Quality Evaluation Protocol} that preserves rule-based answer verification while adding post-hoc LLM-judged diagnostics of useful-step density, low-value step ratios, error localization, and token efficiency.

Across mathematical and scientific benchmarks and two training corpora (DAPO-Math \citep{dapo} and Big-Math \citep{bigmath} ), SCOPE-RL improves accuracy, useful-step density, and token efficiency over outcome-only GRPO on \textbf{Qwen3-8B-Instruct} \citep{qwen3} . Average accuracy rises from 55.80\% to 66.35\% on DAPO-Math and from 53.58\% to 64.79\% on Big-Math, with 16.2\% and 27.1\% token reductions, respectively. The gains hold under a GSPO backend (61.60\%$\to$66.93\%) and on Qwen3-0.6B-Instruct (26.06\%$\to$32.06\%; both verified on DAPO-Math), indicating that reward-signal densification is complementary to policy-update-level RLVR advances.

\paragraph{Contributions.} (i) A unifying \textbf{sparse-anchor} view of RLVR: outcome-only verification leaves the reasoning path under-specified before and after success, calling for phase-specific densification around the same verified endpoint. (ii) \textbf{SCOPE-RL}, instantiating this view with prefix-decomposed verifiable rewards (ASR) and correctness-gated process-shape rewards (QPR) in a naturally sequential procedure. (iii) An expert-validated Step-Quality Evaluation Protocol, with consistent gains across two training sources and multiple benchmarks.

\begin{figure}[t]
\centering
\begin{subfigure}[b]{\linewidth}
\centering
\includegraphics[width=\linewidth]{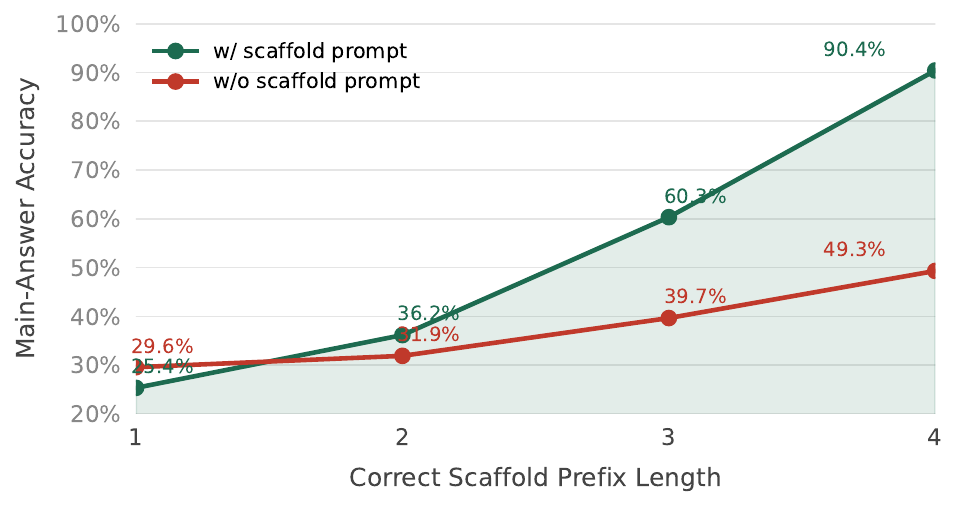}
\caption{Prerequisite scaffold progress vs.\ main-answer accuracy.}
\label{fig:motivation_prefix}
\end{subfigure}

\vspace{0.5em}

\begin{subfigure}[b]{\linewidth}
\centering
\includegraphics[width=\linewidth]{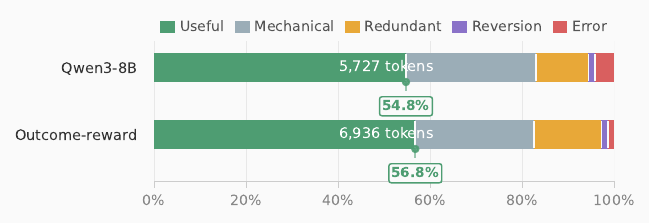}
\caption{Step-quality composition of correct reasoning trajectories.}
\label{fig:motivation_quality}
\end{subfigure}
\caption{Two probes of the same sparse anchor, one on each side of success. \textbf{(a)~Before success}: the same problems are evaluated under a \emph{scaffolded prompt} (sub-questions followed by the main problem) and the \emph{original prompt}, both verified against the same ground-truth answer. Scaffold-prompt accuracy exceeds original-prompt accuracy and rises monotonically with the length of the correct sub-question prefix. \textbf{(b)~After success}: among final-answer-correct responses, outcome-only GRPO produces longer responses that retain substantial low-value content.}
\label{fig:motivation}
\end{figure}

\section{Preliminary}
\label{sec:preliminary}

Given a prompt-answer pair $(q,a)$, outcome-only RLVR samples $y \sim \pi_\theta(\cdot \mid q)$ and assigns the rule-verified reward
\begin{equation}
 r_{\mathrm{out}}(y,a)=\mathbf{1}[\hat{a}(y)=a].
\end{equation}
More generally, GRPO admits any scalar reward $R(q,y)$ in place of $r_{\mathrm{out}}$. For $G$ sampled responses $\{y_g\}_{g=1}^{G}$, we use
\begin{equation}\label{eq:grpo_advantage}
 A_g = \frac{R(q,y_g)-\mathrm{mean}_{j} R(q,y_j)}{\mathrm{std}_{j} R(q,y_j)}.
\end{equation}
When the within-group reward is constant ($\sigma^2_R(q)=0$), Eq.~\eqref{eq:grpo_advantage} is undefined; following standard GRPO implementations~\citep{deepseek-r1}, we set $A_g\equiv 0$, so the group produces no reward-driven policy-gradient signal.

To quantify where GRPO receives usable reward variation, define
\begin{align}
\sigma^2_R(q) &= \mathrm{Var}_g\, R(q,y_g), \label{eq:reward_variance}\\
\eta_R &= \mathbb{P}_{q\sim\mathcal{D}}\!\big[\sigma^2_R(q) > 0\big]. \label{eq:effective_ratio}
\end{align}
$\eta_R$ measures the fraction of prompts that can produce non-zero group-relative advantages under reward $R$; for $r_{\mathrm{out}}\!\in\!\{0,1\}$, $\sigma^2_{r_{\mathrm{out}}}(q)=0$ exactly when the group is uniformly correct or uniformly wrong.

\section{The SCOPE-RL Framework}\label{sec:methods}

\begin{figure*}[t]
\centering
\includegraphics[width=\linewidth]{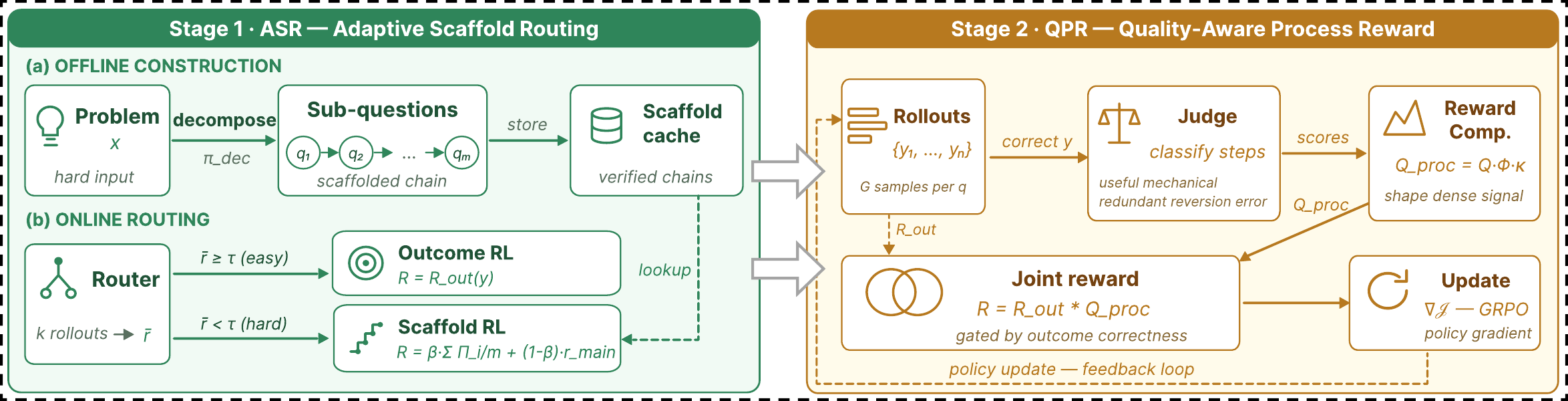}
\caption{Overview of SCOPE-RL. Stage~1 routes hard problems to answer-hidden scaffolds for denser verifiable rewards, while Stage~2 applies correctness-gated process rewards to refine verified-correct trajectories.}
\label{fig:method}
\end{figure*}

\textbf{SCOPE-RL} (\textbf{S}caffolded \textbf{C}hain \textbf{O}ptimization with \textbf{P}rocess \textbf{E}fficiency) instantiates a single design principle on GRPO: the policy update is left unchanged, and the sparse anchor is densified only by modifying \emph{what enters the scalar reward} on each side of the success boundary.

Figure~\ref{fig:method} illustrates the complete pipeline. Both stages follow the same reward-densification template---\textbf{where} new signal is placed, \textbf{when} it is triggered, and \textbf{how} it is composed. Stage~1, \textbf{Adaptive Scaffolded RL} (ASR), places verifiable signal at scaffolded sub-endpoints, triggers it through on-policy routing, and composes prefix-consistent rewards for weak-signal problems. Stage~2, \textbf{Quality-Aware Process RL} (QPR), places signal on already-correct trajectories, triggers it through a correctness gate, and composes bounded process-shape rewards.

\subsection{Two-Stage Coupling}\label{sec:joint_training}

ASR and QPR are coupled by a sequential dependency. Because QPR's correctness gate (\S\ref{sec:stage2}) returns zero on incorrect rollouts, it provides little signal on hard problems before the policy can reliably reach verified answers. ASR first lifts success rates through prefix-decomposed verifiable rewards, after which QPR has a denser set of correct trajectories to shape. Jointly mixing the two rewards would weaken this gate by applying QPR penalties before enough correct rollouts exist to provide useful contrast among them. We therefore train sequentially: Stage~1 produces a checkpoint $\theta_1$ from the routed mixed reward $R_1$, and Stage~2 initializes from $\theta_1$ and optimizes $R_{\mathrm{QPR}}$ on the original undecomposed problems. The only state passed across stages is the policy parameters.

\subsection{Stage 1: Adaptive Scaffolded RL}
\label{sec:stage1}

Stage~1 addresses the before-success regime, where few rollouts reach the final answer and endpoint rewards provide little information about prerequisite progress. Since scaffold-prefix correctness is predictive of main-answer success (Figure~\ref{fig:motivation_prefix}), ASR treats ordered scaffold prefixes as verifiable proxies for prerequisite progress, increasing supervision density through answer-hidden scaffold construction, on-policy scaffold routing, and prefix-consistent scaffold rewards (Figure~\ref{fig:method}, Stage~1).

\subsubsection{Answer-Hidden Scaffold Construction}

Scaffold construction turns hidden prerequisite progress into verifiable sub-endpoints, each checkable by the same rule-based verifier as the final answer. Construction is performed \emph{offline}, with answer-hidden non-leakage enforced as a validity predicate (Eq.~\eqref{eq:scaffold_valid} below). The complete decomposition prompt template is provided in Appendix~\ref{sec:appendix_decompose_prompt}.

Before RL training, ASR generates candidate scaffolded sub-question chains offline for each training problem. Given a problem $q$ with answer $a$, a valid scaffold has the form
\begin{equation}
\mathcal{S}(q) = \bigl\{(q_1, a_1),\, \dots,\, (q_m, a_m),\, (q_{\mathrm{main}}, a)\bigr\},
\end{equation}
where $q_{\mathrm{main}}$ is the final-target sub-problem whose verifiable answer is the original problem's answer $a$, and each $q_i$ targets a prerequisite concept needed for the final solution. During RL training, the policy observes only the ordered sub-questions and the main problem $q_{\mathrm{main}}$; the sub-answers $a_i$ are verifier-only and used solely for reward computation.

We cache a scaffold only when it satisfies
\begin{equation}\label{eq:scaffold_valid}
\mathrm{Valid}(\mathcal{S}) = V(\mathcal{S}) \land P(\mathcal{S}) \land D(\mathcal{S}) \land H(\mathcal{S}),
\end{equation}
where $V$ denotes automatic verifiability, $P$ prerequisite relevance, $D$ dependency awareness, and $H$ answer-hidden non-leakage. These predicates filter out scaffolds that are unverifiable, irrelevant, dependency-inconsistent, or answer-leaking, preserving the verifiable-answer invariant of RLVR (the policy is rewarded only for answers it produces itself, not for answers that appear in the prompt) by construction. A human audit of 200 random scaffolds confirms strict prerequisite-chain integrity (Appendix~\ref{app:scaffold_human_validation}); invalid decompositions are discarded before RL training.

\subsubsection{On-Policy Scaffold Routing}

Applying scaffolds to every problem would dilute supervision and waste rollout cost on already-solvable problems. ASR routes groups by on-policy outcome reward, a self-adapting alternative to static difficulty labels or a learnable router.

During RL, each problem is first rolled out on the original prompt and scored with the outcome reward. The group mean outcome reward
\begin{equation}\label{eq:group_mean_reward}
\bar{r}(q)=\frac{1}{G}\sum_{g=1}^{G} r_{\mathrm{out}}(y_g,a)
\end{equation}
serves as an on-policy estimate of whether endpoint-only training provides sufficient positive signal for $q$. A problem is routed to the scaffolded path if
\begin{equation}
 \bar{r}(q) < \tau,
\end{equation}
where $\tau$ is a fixed routing threshold; otherwise, the original rollouts and outcome rewards are retained. We use $G=8$ and $\tau=0.5$ in all experiments, with sensitivity analysis in Appendix~\ref{app:routing_threshold}.

\subsubsection{Prefix-Consistent Scaffold Rewards}

Once routing has selected a scaffold, the question is how to convert sub-answer matches into reward. We use a prefix product so that a sub-question contributes reward only when itself and all earlier sub-questions in its dependency chain are correct---this stays inside GRPO while preventing the policy from harvesting credit on later sub-answers it can guess without solving the prerequisites.

For routed problems, the scaffolded rollout asks the model to answer all sub-questions and then the original problem in a single generation, using explicit answer tags for extraction. Let $\hat{a}_i$ be the model's answer to $q_i$, and define the prefix-consistency indicator
\begin{equation}
 \Pi_i = \prod_{j=1}^{i} \mathbf{1}[\hat{a}_j = a_j].
\end{equation}
The scaffold reward is
\begin{equation}\label{eq:stage1_reward}
R_{\mathrm{ASR}}
= \beta \sum_{i=1}^{m} \frac{1}{m} \Pi_i
+ (1-\beta) \Pi_m \mathbf{1}[\hat{a}_{\mathrm{main}} = a],
\end{equation}
where $\beta$ controls the weight assigned to scaffold checkpoints; we set $\beta=0.5$ in our experiments. Since $\Pi_m=\prod_{j=1}^{m}\mathbf{1}[\hat a_j=a_j]$ requires all $m$ sub-answers to be correct, valid prerequisite prefixes receive partial credit and the main-answer match contributes only when the entire scaffold prefix is satisfied.

Stage~1 therefore uses the mixed scalar reward
\begin{equation}\label{eq:stage1_mixed_reward}
R_1(q,y)=
\begin{cases}
R_{\mathrm{ASR}}(y,\mathcal{S}(q)), & \bar{r}(q)<\tau, \\
r_{\mathrm{out}}(y,a), & \bar{r}(q)\ge\tau.
\end{cases}
\end{equation}
Here, $y$ denotes the rollout used in the selected branch: scaffolded rollouts for routed groups and original-prompt rollouts otherwise.
\paragraph{Optimizing ASR with GRPO.}
Routed groups are re-rolled out on cached scaffolded prompts and scored with $R_{\mathrm{ASR}}$; non-routed groups keep their original rollouts and outcome rewards.

\paragraph{ASR enlarges the effective gradient support.}
The mixed reward $R_1$ can introduce effective gradient support in routed groups that are degenerate under $r_{\mathrm{out}}$ (Eq.~\eqref{eq:effective_ratio}). In particular, routed scaffold rewards produce non-zero advantages whenever an outcome-uniform group contains prefix-level disagreement, consistent with the gap between GRPO and SCOPE-RL Stage~1 in Figure~\ref{fig:dyn_effective}.

\subsection{Stage 2: Quality-Aware Process RL}
\label{sec:stage2}

Stage~2 addresses the after-success regime, where final-answer rewards cannot distinguish concise, useful, and well-organized correct trajectories from redundant or locally flawed ones. We use \emph{process shape} to denote the compositional structure of an outcome-correct trajectory, including useful-step density, low-value-step ratios, local flaws, and soft conciseness. QPR returns to the original undecomposed problems and applies these preferences only after correctness is established (Figure~\ref{fig:method}, Stage~2).

\subsubsection{Correctness-Gated Process-Shape Reward}

Any process-shape signal that fires on incorrect rollouts could encourage short or fluent wrong answers, reproducing the failure mode of naive length penalties (Figure~\ref{fig:motivation_quality}). QPR prevents this with a multiplicative correctness gate: the rule-based verifier decides final-answer correctness, and the LLM Judge only annotates how a verified-correct trajectory was reached. QPR is initialized from $\theta_1$ and trains on the original, undecomposed problems. Concretely, for each rollout $y$,
\begin{equation}\label{eq:qpr_gate}
R_{\mathrm{QPR}}(y, a) =
\begin{cases}
0, & \hat{a}(y) \neq a, \\
Q_{\mathrm{process}}(y), & \hat{a}(y) = a.
\end{cases}
\end{equation}
The judge model, prompt, and reward formula are fixed throughout QPR training.

\subsubsection{Step-Level Category Annotation}

Conditioned on the gate in Eq.~\eqref{eq:qpr_gate}, QPR uses a fixed LLM Judge to annotate step-level role among correct trajectories. Unlike a learned PRM, the Judge is not trained against the policy and never decides final-answer correctness, so its annotations cannot override the verifier.

For each correct-answer rollout, the judge parses the reasoning trajectory into $N\ge 1$ atomic steps $\{s_1, \dots, s_N\}$. Each step receives a category label $c_i$ from five mutually exclusive classes---\emph{useful}, \emph{mechanical}, \emph{redundant}, \emph{reversion}, and \emph{error}. Only \emph{useful} contributes positive reward, while the remaining four labels carry penalties of different severities. Label definitions, the judging prompt, and parsing details are provided in Appendix~\ref{app:qpr_reward_details}.

From these annotations, we compute the useful-step ratio $S_u$:
\begin{equation}
S_u = \frac{1}{N}\sum_i \mathbf{1}[c_i = u],
\end{equation}
where $u$ denotes the \emph{useful} label. $S_u$ rewards trajectories with a high density of substantive reasoning steps.

\subsubsection{Quality-Aware Reward Composition}

The annotations from the previous section give us a positive useful-step signal and four kinds of low-value step. We compose them with bounded multiplication, which avoids the failure modes of additive composition (length canceling quality) and pure-product composition (a single category zeroing the reward). Concretely, for each undesirable category $k \in \mathcal{K}=\{\mathrm{mec},\mathrm{red},\mathrm{rev},\mathrm{err}\}$, let $r_k$ denote the fraction of steps assigned label $k$, and define
\begin{align}
\Phi &= \prod_{k \in \mathcal{K}} (1 - \lambda_k r_k), \label{eq:qpr_penalty} \\
\kappa &= \frac{1}{1 + \alpha \ln(N+1)}.
\end{align}
The overall process-shape reward is
\begin{equation}\label{eq:qpr_process}
Q_{\mathrm{process}}
= S_u \cdot \Phi \cdot \kappa,
\end{equation}
where $S_u$ rewards useful reasoning density, $\Phi$ suppresses low-value or locally flawed steps, and $\kappa$ softly encourages conciseness with a logarithmic shape that avoids over-penalizing already-short solutions while still discouraging unbounded length. We set all $\lambda_k<1$, so each penalty factor lies in $(0,1]$ and the product $\Phi$ is strictly positive. Since $\kappa$ multiplies the positive term, conciseness alone cannot raise the reward; shortening helps only when $S_u$ is preserved. Hyperparameter values are reported in Appendix~\ref{app:qpr_reward_details}.

\paragraph{Optimizing QPR with GRPO.}
Stage~2 uses $R_{\mathrm{QPR}}$ as the scalar reward in the same GRPO backend. Because QPR keeps all training prompts in their original form, the model must internalize the process-shape preference rather than rely on explicit sub-question structure at inference time.

The complete two-stage training procedure is summarized in Algorithm~\ref{alg:scope_rl} (Appendix~\ref{sec:appendix_algorithm}).

\section{Step-Quality Evaluation Protocol}
\label{sec:step_quality_eval}

We propose a general \textbf{Step-Quality Evaluation Protocol} for diagnosing reasoning processes beyond final-answer accuracy. Accuracy alone cannot reveal whether a training method changes the reasoning trajectories that lead to an answer. The protocol preserves rule-based answer verification while adding post-hoc LLM-judged step annotations, and reports decomposed measures of useful reasoning, low-value steps, error localization, and token efficiency. These metrics are diagnostic-only and never feed back into training. The protocol shares QPR's five-class step taxonomy (\S\ref{sec:stage2}) but uses independent evaluation judges and prompts (Appendix~\ref{app:step_quality_protocol}). We instantiate it on standard scientific and mathematical reasoning benchmarks; a scaling validation across five model sizes confirms the protocol's discriminative power (Appendix~\ref{sec:appendix_scaling}).

\begin{table*}[htbp]
\centering
\small
\setlength{\tabcolsep}{4pt}
\resizebox{\textwidth}{!}{%
\begin{tabular}{l ccccc cc r}
\toprule
\multirow{2}{*}{\textbf{Method}}
  & \multicolumn{5}{c}{\textbf{Accuracy (\%) $\uparrow$}}
  & \multicolumn{2}{c}{\textbf{Reasoning Quality $\uparrow$}}
  & \multirow{2}{*}{\textbf{Avg.\ Tokens $\downarrow$}} \\
\cmidrule(lr){2-6}\cmidrule(lr){7-8}
& GPQA\textsubscript{@1} & MATH500\textsubscript{@1} & AIME24\textsubscript{@8} & AIME25\textsubscript{@8} & Avg & Useful\% & 1st-Err-Pos & \\
\midrule
Qwen3-8B-Instruct & 46.97 & 69.92 & 26.25 & 18.75 & 40.47 & 54.76 & 0.665 & 5{,}727 \\
\midrule
\multicolumn{9}{l}{\textit{Trained on DAPO-Math}} \\
\midrule
GRPO               & 48.48 & 79.70 & 51.67 & 43.33 & 55.80 & 56.79 & 0.606 & 6{,}936 \\
\addlinespace[3pt]
ASR only            & 51.52 & 90.23 & 65.00 & \textbf{56.67} & 65.86 & 60.73 & 0.638 & 6{,}948 \\
SCOPE-RL            & \textbf{53.03} & \textbf{90.30} & \textbf{65.83} & 56.25 & \textbf{66.35} & \textbf{64.52} & \textbf{0.703} & \textbf{5{,}815} \\
\midrule
\multicolumn{9}{l}{\textit{Trained on Big-Math (data-source generalisation)}} \\
\midrule
GRPO               & 43.43 & 84.21 & 46.67 & 40.00 & 53.58 & 59.73 & 0.656 & 8{,}228 \\
\addlinespace[3pt]
ASR only            & 53.03 & \textbf{87.22} & 59.17 & 50.00 & 62.36 & 61.92 & 0.666 & 6{,}245 \\
SCOPE-RL            & \textbf{55.05} & 84.96 & \textbf{64.58} & \textbf{54.58} & \textbf{64.79} & \textbf{66.47} & \textbf{0.738} & \textbf{5{,}994} \\
\bottomrule
\end{tabular}
}
\caption{Main results on accuracy, reasoning quality, and efficiency. All methods use GRPO; ASR denotes SCOPE-RL after Stage~1, before QPR.
\textbf{Accuracy}: subset accuracy and arithmetic mean (Avg), with truncated responses counted incorrect.
\textbf{Useful\%}: fraction of LLM-judge-labeled useful steps.
\textbf{1st-Err-Pos}: relative position of the first error step ($\uparrow$ = later errors).
\textbf{Avg.\ Tokens}: mean tokens per sample.
Bold marks the best result within each training-data group.}
\label{tab:main_results}
\end{table*}

\section{Experiments}
\label{sec:experiments}

We evaluate SCOPE-RL in a math/science-oriented RLVR setting with Qwen-family backbones, testing whether its stage-specific supervision---ASR before success and QPR after success---improves \textbf{learnability}, \textbf{trace quality}, and \textbf{efficiency}, and whether these gains hold across two training sources, four evaluation domains, and independent expert preferences.

\paragraph{Setup.}
We use Qwen3-8B-Instruct \citep{qwen3} as the base model and outcome-only GRPO \citep{grpo} as the primary baseline, trained for the same total optimizer budget as the two SCOPE-RL stages combined. SCOPE-RL is trained sequentially on 2{,}400 problems from each of DAPO-Math \citep{dapo} and Big-Math \citep{bigmath}: Stage~1 applies ASR; its best checkpoint initializes Stage~2 QPR. We evaluate GPQA \citep{gpqa} and MATH500 \citep{lightman2023} with single-sample decoding, and report 8-sample average accuracy on AIME 2024 \citep{aime24} and AIME 2025 \citep{aime25}. Full training, fairness, and evaluation-protocol details are in Appendix~\ref{app:training_details}.

\subsection{Main Results and Reasoning Efficiency}
\label{sec:main_results}

Across both training sources, ASR raises final-answer accuracy over outcome-only GRPO; QPR then improves correct-trace quality and efficiency without sacrificing correctness (Table~\ref{tab:main_results}).

\paragraph{Accuracy.}
ASR raises average accuracy on both sources (DAPO-Math: 55.80$\to$65.86\%; Big-Math: 53.58$\to$62.36\%), with the largest gains on AIME24/25 where endpoint rewards are sparsest. QPR adds further headroom (66.35\% / 64.79\%) and improves process diagnostics: Useful\% rises monotonically across the two stages, and 1st-Err-Pos shifts later. Pass@128 analysis (Appendix~\ref{sec:appendix_passk}) rules out a capacity ceiling from ASR scaffolding; QPR largely preserves this ceiling, with a minor coverage--quality trade on one AIME subset.

\paragraph{Efficiency.}
Total token reduction is 16.2\% (DAPO-Math) and 27.1\% (Big-Math) without sacrificing accuracy or useful-step density. The split is source-dependent: on DAPO-Math, Stage~1 preserves length (6{,}936$\to$6{,}948) and Stage~2 carries the full reduction; on Big-Math, Stage~1 already cuts 24.1\% (8{,}228$\to$6{,}245) and Stage~2 refines a further 4.0\%. QPR thus shifts from primary compressor to fine-grained refiner with Stage~1's residual redundancy.

\subsection{Ablation Studies}
\label{sec:ablation}

Our ablations test whether the gains depend on prefix-decomposed verifiable rewards, correctness-gated process-shape rewards, and sequential training rather than auxiliary prompting, static routing, or simple brevity pressure.

\paragraph{ASR needs both scaffolded rewards and adaptive routing.}
Removing scaffolded decomposition (No Decomp.), applying it indiscriminately (All Decomp.), or replacing on-policy routing with offline difficulty labels each costs 7--9~pp of average accuracy (Table~\ref{tab:ablation_stage1}); both prefix-decomposed rewards and capability-tracking routing matter. A reward-signal ablation (Appendix~\ref{app:asr_reward_ablation}) isolates the prefix-consistency gate as the largest contributor.

\begin{table}[t]
\centering
\scriptsize
\setlength{\tabcolsep}{2pt}
\resizebox{\columnwidth}{!}{
\begin{tabular}{lccccc}
\toprule
\textbf{Method} & \textbf{GPQA\textsubscript{@1}} & \textbf{M500\textsubscript{@1}} & \textbf{A24\textsubscript{@8}} & \textbf{A25\textsubscript{@8}} & \textbf{Avg} \\
\midrule
GRPO                & 48.48 & 79.70 & 51.67 & 43.33 & 55.80 \\
No Decomp.          & 45.45 & 82.09 & 54.58 & 45.42 & 56.89 \\
All Decomp.         & 46.97 & 77.61 & 59.58 & 50.00 & 58.54 \\
Offline Route       & 42.42 & 84.33 & 57.50 & 51.25 & 58.88 \\
\textbf{ASR only}    & \textbf{51.52} & \textbf{90.23} & \textbf{65.00} & \textbf{56.67} & \textbf{65.86} \\
\bottomrule
\end{tabular}
}
\caption{ASR (Stage~1) ablation. \textit{No Decomp.}: drop scaffolded rewards, keep dynamic sampler; \textit{All Decomp.}: scaffold all problems, no routing; \textit{Offline Route}: fixed difficulty labels instead of on-policy routing; \textit{ASR only}: full Stage~1 model. M500: MATH500; A24/A25: AIME 2024/2025.}

\label{tab:ablation_stage1}
\end{table}


\begin{table}[t]
\centering
\small
\setlength{\tabcolsep}{4pt}
\resizebox{\columnwidth}{!}{%
\begin{tabular}{l cccr}
\toprule
\textbf{Method} & \textbf{Avg} & \textbf{Useful\%} & \textbf{1st-Err} & \textbf{Tokens} \\
\midrule
\multicolumn{5}{l}{\textit{(a) Simple alternatives}} \\
\midrule
Holistic Judge        & 63.98 & 64.14 & 0.707 & 7{,}277 \\
Concise Prompt        & 62.53 & 58.69 & 0.662 & \textbf{4{,}115} \\
Overlong Penalty      & 59.81 & 58.97 & 0.708 & 6{,}424 \\
\midrule
\multicolumn{5}{l}{\textit{(b) Reward-component ablations}} \\
\midrule
No Positive           & 58.81 & 58.84 & 0.658 & 4{,}281 \\
No Penalty            & 60.86 & 61.00 & \textbf{0.751} & 5{,}328 \\
No Length             & 63.33 & 62.37 & 0.700 & 7{,}882 \\
\midrule
\multicolumn{5}{l}{\textit{(c) Training-schedule ablation}} \\
\midrule
Mix-Stage             & 60.98 & \textbf{65.86} & 0.727 & 6{,}518 \\
\midrule
\textbf{SCOPE-RL}     & \textbf{66.35} & 64.52 & 0.703 & 5{,}815 \\
\bottomrule
\end{tabular}
}
\caption{QPR (Stage~2) ablations on the shared ASR Stage~1 checkpoint; panel~(c) \textit{Mix-Stage} is the no-warm-up control with mixed ASR/QPR rewards. \textit{Concise Prompt}: inference-time instruction on the same checkpoint, no QPR training (Appendix~\ref{sec:appendix_concise_prompt}).}

\label{tab:qpr_ablation_panels}
\end{table}

\paragraph{Step-level decomposition, all three terms, and sequential staging are each load-bearing.}
Table~\ref{tab:qpr_ablation_panels} reports three coupled ablations on the shared ASR backbone. Panel~(a): step-level decomposition is the load-bearing choice---\textit{Holistic Judge}, which reuses the same correctness gate and LLM judge but collapses per-step labels into a single trace score, nearly matches SCOPE-RL on Useful\% (64.14\%) yet trails 2.37~pp on Avg and uses 25\% more tokens, while \textit{Concise Prompt} sacrifices useful-step density (58.69\%) and \textit{Overlong Penalty} weakens accuracy without reliably controlling length. Panel~(b): the three multiplicative terms in $Q_{\mathrm{process}}$ (Eq.~\eqref{eq:qpr_process}) target disjoint failure modes---removing the positive term collapses Avg (58.81\%), removing the penalty leaves redundancy (Useful\% 61.00\%) despite the latest 1st-Err position (0.751), and removing the length factor preserves accuracy but inflates tokens to 7{,}882. Panel~(c): \textit{Mix-Stage} reaches only 60.98\% Avg, confirming that prefix-decomposed verifiable rewards must precede process-shape refinement. Although individual ablations lead on isolated metrics (No Penalty on 1st-Err, Mix-Stage on Useful\%), only SCOPE-RL jointly maximizes accuracy and overall trace quality.

\subsection{Training Dynamics}
\label{sec:dynamics}

\begin{figure*}[htbp]
\centering
\begin{minipage}[t]{0.32\textwidth}
    \centering
    \includegraphics[width=\linewidth]{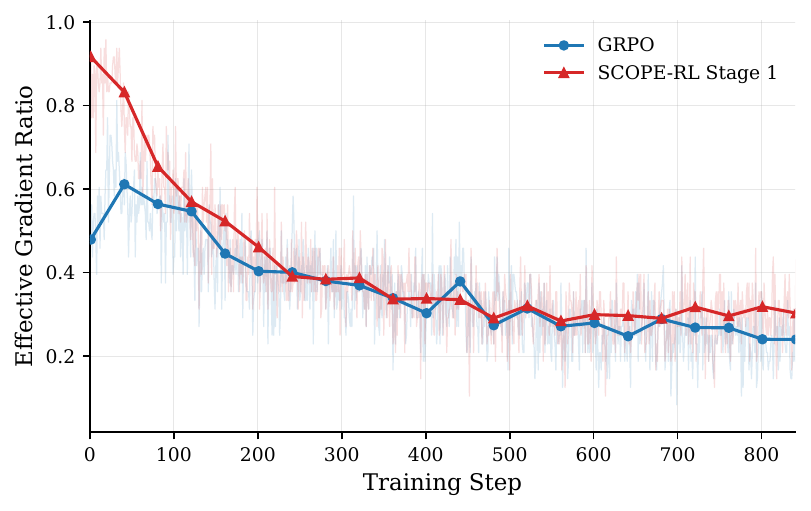}
    \subcaption{Effective gradient ratio}
    \label{fig:dyn_effective}
\end{minipage}
\hfill
\begin{minipage}[t]{0.32\textwidth}
    \centering
    \includegraphics[width=\linewidth]{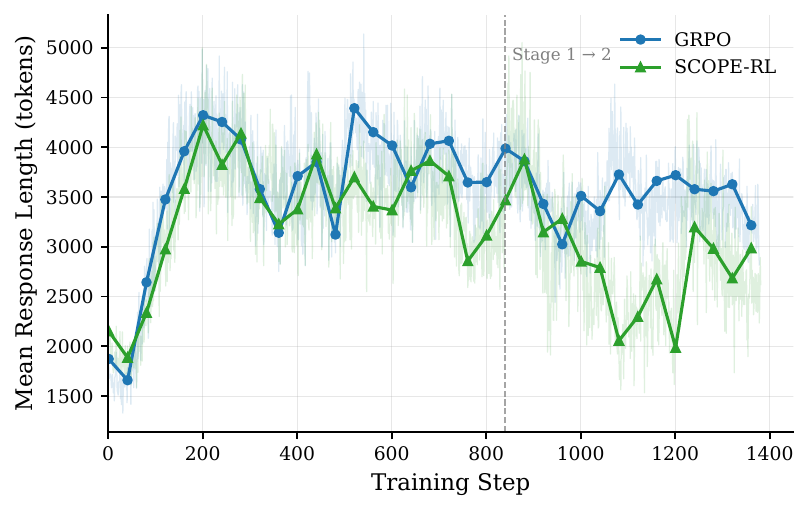}
    \subcaption{Mean response length}
    \label{fig:dyn_length}
\end{minipage}
\hfill
\begin{minipage}[t]{0.32\textwidth}
    \centering
    \includegraphics[width=\linewidth]{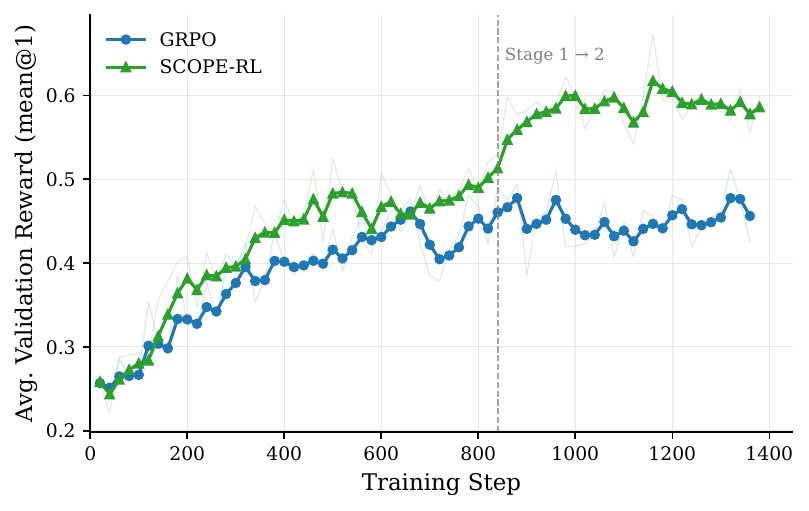}
    \subcaption{Avg.\ validation reward}
    \label{fig:dyn_reward}
\end{minipage}
\caption{Training dynamics of GRPO vs.\ SCOPE-RL. (a)~ASR maintains a higher effective gradient ratio via scaffolded routing. (b)~Response length grows in Stage~1 then drops in Stage~2 as QPR encourages conciseness. (c)~Validation reward improves across both stages. Dashed line marks the stage transition.}
\label{fig:training_dynamics}
\end{figure*}

Figure~\ref{fig:training_dynamics} corroborates the two-stage mechanism: ASR sustains a higher \emph{effective gradient ratio}---the fraction of groups with non-degenerate advantages---by routing low-pass-rate problems to scaffolded prompts; response length grows under ASR then drops once QPR engages, while validation reward rises, indicating that QPR removes low-value reasoning rather than necessary steps.

\subsection{Qualitative Analysis}

\begin{figure}[t]
\centering
\includegraphics[width=\linewidth]{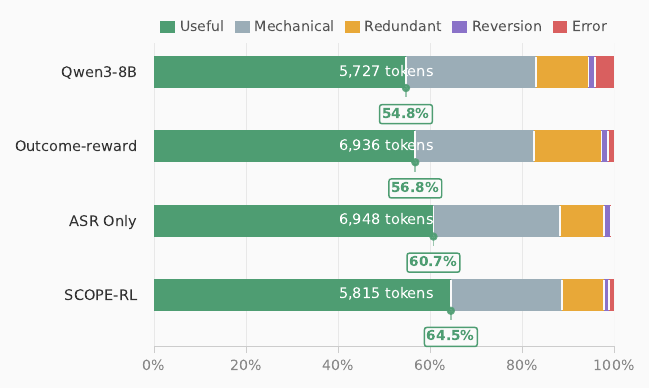}
\caption{Step-quality composition of correct traces across training stages. Bars show proportions of useful, mechanical, redundant, reversion, and error steps, with mean token count and useful-step percentage.}
\label{fig:step_composition}
\end{figure}

Figure~\ref{fig:step_composition} decomposes correct traces into the five-category taxonomy. The progression matches Table~\ref{tab:main_results}: GRPO inflates length without raising useful-step density, ASR shifts steps toward useful reasoning, and QPR drops length while pushing Useful\% to its peak.

A side-by-side trace comparison on the same Lagrange interpolation problem (Appendix~\ref{sec:appendix_case_study}) makes the two-stage mechanism concrete: GRPO reaches an incorrect intermediate conclusion and recovers only via lengthy self-correction, ASR identifies the correct mathematical structure earlier though some redundant verification remains, and QPR distills the trace into a direct partition-of-unity derivation.

\subsection{Robustness Across Optimizer Backend and Model Scale}
\label{sec:robustness}

SCOPE-RL's gains hold under a GSPO backend (61.60$\to$66.93\%, 13.1\% token reduction) and on Qwen3-0.6B-Instruct (26.06$\to$32.06\%, 22.9\% token reduction), indicating that the before/after-success decomposition is not tied to a single optimizer or model scale. Full results are in Appendix~\ref{app:robustness_results}.

\subsection{Additional Process-Quality Validation}
\label{sec:analysis_and_cases}

We further validate that QPR improves perceived trace quality rather than only automatic diagnostic scores. In a pairwise evaluation on 200 examples where both GRPO and SCOPE-RL are answer-correct, three domain experts prefer SCOPE-RL across all five dimensions: conciseness (72\% vs.\ 28\%), non-redundancy (69\% vs.\ 31\%), clarity (65.5\% vs.\ 34.5\%), logical coherence (59.5\% vs.\ 40.5\%), and overall quality (70.5\% vs.\ 29.5\%). Full results are reported in Appendix~\ref{app:llm_judge_eval}.

\section{Related Work}

\noindent\textbf{RLVR and hard-problem learning.}
Recent RLVR systems and GRPO-style optimizers \citep{deepseek-r1, grpo, dapo, gspo, drgrpo} improve LLM reasoning under verifiable final-answer rewards but face sparse signal on hard problems, motivating curricula \citep{e2h, reasoningcurriculum, adarft}, difficulty filtering \citep{rorl}, demonstrations \citep{luffy}, hint or prefix injection \citep{pope, sage, evocot, stephint}, reformulation or decomposition \citep{cogdrift, ladder}, and staged pipelines \citep{relift}. These methods reshape inputs, rollouts, or GRPO updates \citep{stephint, luffy}, but none install independently verifiable sub-answer rewards. ASR instead places \emph{rule-based verifiers on each scaffolded sub-answer}, hidden from the policy. EvoCoT \citep{evocot} conditions rollouts on self-generated, verified CoT prefixes progressively shortened as a curriculum, not on independently verified sub-answer targets.

\noindent\textbf{Process supervision and reasoning efficiency.}
Step-level supervision spans PRMs trained on step labels \citep{lightman2023, mathshepherd}, implicit process rewards from policy logits \citep{cui2025process}, refined aggregation or credit assignment of PRM rewards \citep{pure, progrs}, and analytical decomposition of the outcome objective \citep{prl}; PRIME \citep{prime} supports verifier evaluation and selection rather than step labels, and efficiency methods reduce overthinking via length-aware or compression objectives \citep{thinkprune, smartthinker, ccc}. QPR differs in how it gates and composes process signal: conditional on a correct answer, a fixed (not learned) LLM judge labels each step's \emph{functional role} over a closed taxonomy (useful, mechanical, redundant, reversion, error) and shapes the trajectory reward accordingly, with zero process reward on incorrect rollouts. QPR operates on the reward-signal axis while ASR exposes additional verifiable targets through answer-hidden scaffolds; both leave the policy-update rule unchanged.

\section{Conclusion}
We framed outcome-only RLVR as resting on a \emph{sparse anchor}---reliable at the endpoint but under-specified on the reasoning path before and after success. \textbf{SCOPE-RL} densifies this anchor while retaining the GRPO update: \textbf{Adaptive Scaffolded RL} adds prefix-decomposed verifiable rewards on answer-hidden scaffolded sub-question chains before success, and \textbf{Quality-Aware Process RL} applies correctness-gated process-shape rewards to refine verified-correct trajectories after success. Together with the Step-Quality Evaluation Protocol, experiments on Qwen3-8B-Instruct show gains in accuracy, token efficiency, and process quality across DAPO-Math and Big-Math, with robustness checks under GSPO and on Qwen3-0.6B-Instruct suggesting that reward-signal densification is complementary to policy-update-level RLVR advances.

\section*{Limitations}
Our experiments are limited to Qwen-family backbones and math/science-oriented RLVR settings. The main results use Qwen3-8B-Instruct, with additional validation on Qwen3-0.6B-Instruct and a GSPO backend, but do not cover other model families, larger-scale models, or broader verifiable-reward domains.

We do not exhaustively compare against all hint-based exploration or length-aware reasoning-compression methods. This is a deliberate scope choice: our experiments isolate whether verifier-compatible reward densification improves RLVR when the policy optimizer and training interface are fixed. Combining SCOPE-RL with these orthogonal techniques is a promising direction for future work.

Finally, our hyperparameter analysis is limited to a sensitivity sweep for the routing threshold $\tau$ (Appendix~\ref{app:routing_threshold}); broader tuning of $\beta$, $\alpha$, and $\{\lambda_k\}$ remains future work.

\bibliography{custom}

\clearpage
\appendix

\section{Training Algorithm}
\label{sec:appendix_algorithm}

Algorithm~\ref{alg:scope_rl} presents the complete pseudocode for the SCOPE-RL two-stage training procedure described in Section~\ref{sec:methods}. 

\begin{algorithm}[H]
\caption{SCOPE-RL: Two-Stage Training Framework}
\label{alg:scope_rl}
\renewcommand{\algorithmicrequire}{\textbf{Input:}}
\begin{algorithmic}[1]
\REQUIRE Base model $\pi_\theta$, training data $\mathcal{D}$, scaffold function $\mathcal{S}(\cdot)$, threshold $\tau$, group size $G$, LLM Judge $\mathcal{J}$
\STATE \textbf{--- Stage 1: Adaptive Scaffolded RL ---}
\FOR{each training step}
    \STATE Sample prompt batch $\{q_1, \dots, q_B\}$ from $\mathcal{D}$
    \FOR{each prompt $q_j$}
        \STATE Generate $G$ rollouts from $\pi_\theta$ on $q_j$; compute $\bar{r}(q_j)$
        \IF{$\bar{r}(q_j) < \tau$}
            \STATE Replace $q_j$ with scaffolded prompt $\mathcal{S}(q_j)$; re-generate $G$ rollouts
            \STATE Compute $R_{\mathrm{ASR}}$ via Eq.~\eqref{eq:stage1_reward}
        \ELSE
            \STATE Retain original rollouts with outcome reward $r_{\mathrm{out}}$
        \ENDIF
    \ENDFOR
    \STATE Merge all groups into unified batch; update $\pi_\theta$ with the chosen RLVR optimizer
\ENDFOR
\STATE \textbf{--- Stage 2: Quality-Aware Process RL ---}
\STATE Initialize $\pi_\theta$ from best Stage~1 checkpoint
\FOR{each training step}
    \STATE Sample prompt batch; generate $G$ rollouts from $\pi_\theta$ on original prompts
    \FOR{each rollout $y$}
        \IF{$\hat{a}(y) = a$}
            \STATE Invoke $\mathcal{J}$ to annotate steps with category labels
            \STATE Compute $Q_{\mathrm{process}}(y)$ via Eq.~\eqref{eq:qpr_process}; set $R_{\mathrm{QPR}}=Q_{\mathrm{process}}(y)$
        \ELSE
            \STATE Set $R_{\mathrm{QPR}}=0$
        \ENDIF
    \ENDFOR
    \STATE Update $\pi_\theta$ with the chosen RLVR optimizer using $R_{\mathrm{QPR}}$ as reward
\ENDFOR
\end{algorithmic}
\end{algorithm}

\section{Pass@k Analysis}
\label{sec:appendix_passk}

Since ASR employs an offline teacher model to generate scaffolded decompositions for difficult problems, a natural concern is whether this auxiliary supervision constrains the model's exploration capacity. We address this with a pass@$k$ analysis on AIME 2024 \citep{aime24}, AIME 2025-I, and AIME 2025-II \citep{aime25}; we read pass@$k{=}128$ as an upper-bound proxy for reasoning capacity, since at this budget any solution the model can in principle find is likely to surface in at least one of 128 samples.

At $k{=}128$, ASR matches or exceeds GRPO on every subset (AIME24: 90.00\% vs.\ 90.00\%; AIME25-I: 93.33\% vs.\ 86.67\%; AIME25-II: 93.33\% vs.\ 86.67\%) and exceeds the base model by 17--27~pp. Scaffolded training therefore does not cap reasoning potential---if anything, it raises the ceiling on the harder AIME 2025 subsets.

SCOPE-RL retains the GRPO pass@128 capacity on AIME 2024 (90.00\%) and AIME 2025-II (86.67\%), but drops on AIME 2025-I (86.67\%$\to$80.00\%). This is consistent with QPR's design as a quality refiner rather than a capacity expander: by concentrating probability mass on cleaner, shorter derivations, the policy trades a small amount of solution diversity for higher useful-step density and lower token cost (Tab.~\ref{tab:main_results}). Capacity itself is established at ASR; SCOPE-RL then operates on top of that capacity. For applications that prioritize raw coverage at large sampling budgets, the ASR checkpoint is the appropriate operating point; SCOPE-RL is shipped because the deployment regime targets accuracy and efficiency at small $k$ rather than the high-$k$ frontier.

\begin{table*}[tp]
\centering
{\setlength{\tabcolsep}{12pt}%
\resizebox{\textwidth}{!}{%
\begin{tabular}{@{\hskip 1.5em} ll cccccc @{\hskip 1.5em}}
\toprule
\textbf{Dataset} & \textbf{Method} & $k{=}1$ & $k{=}8$ & $k{=}16$ & $k{=}32$ & $k{=}64$ & $k{=}128$ \\
\midrule
\multirow{4}{*}{AIME 2024}
& Qwen3-8B-Instruct & 26.67 & 46.67 & 60.00 & 66.67 & 66.67 & 66.67 \\
& GRPO       & 60.00 & 80.00 & 83.33 & 86.67 & 90.00 & 90.00 \\
& ASR only    & 66.67 & 83.33 & 83.33 & 83.33 & 86.67 & 90.00 \\
& SCOPE-RL    & 63.33 & 83.33 & 83.33 & 83.33 & 86.67 & 90.00 \\
\midrule
\multirow{4}{*}{AIME 2025-I}
& Qwen3-8B-Instruct & 26.67 & 46.67 & 53.33 & 53.33 & 66.67 & 66.67 \\
& GRPO       & 40.00 & 66.67 & 66.67 & 66.67 & 86.67 & 86.67 \\
& ASR only    & 53.33 & 66.67 & 73.33 & 80.00 & 93.33 & 93.33 \\
& SCOPE-RL    & 53.33 & 73.33 & 80.00 & 80.00 & 80.00 & 80.00 \\
\midrule
\multirow{4}{*}{AIME 2025-II}
& Qwen3-8B-Instruct & 20.00 & 26.67 & 40.00 & 53.33 & 60.00 & 73.33 \\
& GRPO       & 46.67 & 86.67 & 86.67 & 86.67 & 86.67 & 86.67 \\
& ASR only    & 60.00 & 86.67 & 86.67 & 86.67 & 93.33 & 93.33 \\
& SCOPE-RL    & 53.33 & 86.67 & 86.67 & 86.67 & 86.67 & 86.67 \\
\bottomrule
\end{tabular}
}
}
\caption{Pass@$k$ accuracy (\%) on AIME subsets. Reasoning capacity is read at $k{=}128$: ASR matches or exceeds GRPO on all three subsets, showing scaffolded training does not cap potential. SCOPE-RL preserves the GRPO capacity on two of three subsets and trades coverage for trace quality on AIME 2025-I, consistent with QPR's role as a quality refiner rather than a capacity expander.}
\label{tab:passk}
\end{table*}

\section{Motivation Probing Experiment Details}
\label{app:motivation_probe}

To verify that scaffold-prefix progress reflects genuine path-level learning signal rather than merely correlating with intrinsic problem easiness, we design the following probing experiment.

\paragraph{Data construction.}
We sample 755 problems from the training set whose scaffold decompositions contain exactly 4 sub-questions. Fixing the number of sub-questions eliminates chain-length variation as a confound in the binning analysis.

\paragraph{Evaluation procedure.}
We evaluate the base model (Qwen3-8B-Instruct \citep{qwen3}) on these 755 problems under two conditions: (1)~\emph{Scaffold prompt}: the model receives the full sub-question chain together with the original main problem and is asked to answer all sub-questions and the main problem sequentially; (2)~\emph{Original prompt}: the model receives only the original, undecomposed problem.

\paragraph{Binning and metric computation.}
From the scaffold-prompt rollouts, we determine the longest correct prefix for each sample---i.e., the number of consecutive sub-questions answered correctly before the first failure (ranging from 1 to 4). Samples are grouped into four bins by this prefix length. For each bin, we report two main-answer accuracy values: (a)~accuracy under the scaffold prompt (green line in Figure~\ref{fig:motivation_prefix}), and (b)~accuracy when the same set of problems is evaluated under the original prompt (red line in Figure~\ref{fig:motivation_prefix}). The comparison between the two lines on the same bin isolates the effect of scaffold-provided structure from problem-level difficulty.

\section{Reward Properties: Bounded Scale and Effective Gradient Support}
\label{app:reward_bounds}

We analyze two properties of the stage-specific rewards. \textbf{Bounded scale} (\S\ref{app:bounded_scale}) keeps both rewards on the same $[0,1]$ range as $r_{\mathrm{out}}$, so the GRPO clip range and KL coefficient need not be re-tuned per stage. \textbf{Effective gradient support} (\S\ref{app:effective_support}) shows that bounded scale alone does not ensure learning: on the prompt subsets where outcome-only training is degenerate ($\sigma^2_{r_{\mathrm{out}}}=0$), ASR and QPR can each provide \emph{strictly positive} reward variance under the heterogeneity conditions stated below.

\subsection{Bounded Scale}
\label{app:bounded_scale}

For ASR, with $m\ge1$, $\beta\in[0,1]$, $\Pi_i\in\{0,1\}$, and $\mathbf{1}[\hat a_{\mathrm{main}}=a]\le1$, Eq.~\eqref{eq:stage1_reward} gives
\begin{equation}
0 \le R_{\mathrm{ASR}}
\le \beta + (1-\beta) = 1.
\end{equation}

For QPR, with $S_u,r_k\in[0,1]$, $\lambda_k\in[0,1)$, and $\alpha\ge0$, each penalty factor satisfies $(1-\lambda_k r_k)\in(0,1]$, so $\Phi\in(0,1]$; also $\kappa=1/(1+\alpha\ln(N+1))\in(0,1]$ for $N\ge1$. Eq.~\eqref{eq:qpr_process} gives
\begin{equation}
0 \le Q_{\mathrm{process}}
\le S_u
\le 1.
\end{equation}
Since $R_{\mathrm{QPR}}=Q_{\mathrm{process}}$ for verifier-correct rollouts and $R_{\mathrm{QPR}}=0$ otherwise, $0\le R_{\mathrm{QPR}}\le1$ follows.

\subsection{Effective Gradient Support}
\label{app:effective_support}

A bounded reward stays on a learnable scale, but GRPO requires within-group reward variation to produce non-zero advantages. Concretely, the std-zero convention (\S\ref{sec:preliminary}) makes the per-rollout advantage piecewise:
\begin{equation}
A_g(q) =
\begin{cases}
\dfrac{R(q,y_g)-\bar R(q)}{\sigma_R(q)}, & \sigma^2_R(q) > 0, \\[0.4em]
0, & \sigma^2_R(q) = 0,
\end{cases}
\label{eq:advantage_piecewise}
\end{equation}
where $\bar R(q)=\mathrm{mean}_j R(q,y_j)$ and $\sigma_R(q)=\sqrt{\sigma^2_R(q)}$.
Throughout this appendix, $\mathrm{Var}_g$ denotes the empirical population variance over the $G$ sampled rollouts. The policy receives reward-driven gradient on $q$ if and only if $\sigma^2_R(q)>0$. The effective gradient ratio $\eta_R = \mathbb{P}_q[\sigma^2_R(q)>0]$ (Eq.~\eqref{eq:effective_ratio}) measures the fraction of prompts that pass this threshold.

For the outcome-only reward $r_{\mathrm{out}}\!\in\!\{0,1\}$, let $p(q) = \frac{1}{G}\sum_g r_{\mathrm{out}}(y_g,a)$ denote the empirical pass rate in the sampled group. Then
\begin{equation}\label{eq:rout_variance}
\sigma^2_{r_{\mathrm{out}}}(q) = p(q)\bigl(1-p(q)\bigr),
\end{equation}
which vanishes exactly when $p(q)\in\{0,1\}$, i.e.\ when the group is uniformly correct or uniformly wrong. The fraction of prompts contributing reward-driven gradient under outcome-only training is therefore $\eta_{r_{\mathrm{out}}}=\mathbb{P}_q[0<p(q)<1]$, the prompts where the current policy is neither saturated nor fully blocked.

\paragraph{ASR recovers signal via prefix disagreement.}
Let $L(y_g)=\sum_{i=1}^m \Pi_i(y_g)$ denote the length of the correct scaffold prefix. Since the prefix indicators are nested ($\Pi_i$ requires consecutive correctness on sub-questions $1$ through $i$, so $\Pi_1\!\ge\!\Pi_2\!\ge\!\cdots\!\ge\!\Pi_m$), $\Pi_m=\mathbf{1}[L(y_g)=m]$, and Eq.~\eqref{eq:stage1_reward} rewrites as
\begin{equation}\label{eq:asr_via_L}
\begin{aligned}
R_{\mathrm{ASR}}(y_g) ={}& \tfrac{\beta}{m}\,L(y_g) \\
&{}+ (1{-}\beta)\,\mathbf{1}[L(y_g){=}m]\,\mathbf{1}[\hat a_{\mathrm{main}}{=}a].
\end{aligned}
\end{equation}
If $L(y_g)$ differs across rollouts in a routed group and $\beta>0$, then $R_{\mathrm{ASR}}$ takes distinct values across rollouts, so $\sigma^2_{R_{\mathrm{ASR}}}(q)>0$. This holds in particular on routed groups with $\sigma^2_{r_{\mathrm{out}}}(q)=0$: ASR recovers gradient signal on this degenerate subset whenever scaffold-prefix lengths disagree across rollouts, without requiring any assumption on the main-answer distribution under the scaffolded re-rollouts.

\textbf{Remark on $\Pi_m$ as a verification gate.} The factor $\Pi_m$ in the second term means that when $\Pi_m=0$ on all rollouts, the main-answer match contributes nothing even if some $\hat a_{\mathrm{main}}(y_g)=a$ holds by coincidence. By design, this prevents $R_{\mathrm{ASR}}$ from rewarding lucky main-answer guesses without prerequisite progress, in line with ASR's design intent of crediting verifiable progress rather than coincidence; on such ``lucky'' rollouts $r_{\mathrm{out}}$ would have variance, but the second term of $R_{\mathrm{ASR}}$ intentionally contributes none. The empirical gap between GRPO and SCOPE-RL Stage~1 in Figure~\ref{fig:dyn_effective} indicates that ASR's prefix signal outweighs the lucky-guess variance suppressed by $\Pi_m$ in practice.

\paragraph{QPR recovers signal on outcome-uniform-correct groups.}
Let $p_c(q)=\frac{1}{G}\sum_g \mathbf{1}[\hat a(y_g)=a]$ denote the correct-rollout fraction in the sampled group (equivalently, $p_c\equiv p$). If $p_c(q)=0$, then $R_{\mathrm{QPR}}$ is identically zero and $\sigma^2_{R_{\mathrm{QPR}}}(q)=0$. For $p_c(q)>0$, let $\mu_c, \nu_c$ denote the mean and variance of $Q_{\mathrm{process}}(y)$ conditioned on $\hat a(y)=a$. The QPR gate (Eq.~\eqref{eq:qpr_gate}) makes $R_{\mathrm{QPR}}$ a mixture of $\{0\}$ and $Q_{\mathrm{process}}$, and a law-of-total-variance decomposition gives
\begin{equation}\label{eq:qpr_var_decomp}
\sigma^2_{R_{\mathrm{QPR}}}(q) = p_c\,\nu_c + p_c(1-p_c)\,\mu_c^2,
\end{equation}
where $p_c, \mu_c, \nu_c$ are evaluated at $q$ and $\mu_c\ge 0$ since $Q_{\mathrm{process}}\in[0,1]$.
Eq.~\eqref{eq:qpr_var_decomp} is strictly positive iff $p_c(q)>0$ and either (i) $\nu_c(q)>0$ (heterogeneity of $Q_{\mathrm{process}}$ among correct rollouts), or (ii) $0<p_c(q)<1$ and $\mu_c(q)>0$ (mixed group with a non-zero typical correct-rollout reward).
The first term captures intra-correct variance in process shape; the second is the binary-pass-rate variance scaled by $\mu_c^2$. On groups with mixed correct and incorrect rollouts ($0<p_c<1$), $\sigma^2_{r_{\mathrm{out}}}$ is already positive; QPR additionally provides finer-grained variation under condition (i) or (ii), and loses signal only in the boundary case where the judge returns identically zero on all correct rollouts ($\mu_c=\nu_c=0$). The genuine recovery happens on \emph{outcome-uniform-correct} groups ($p_c=1$, where $\sigma^2_{r_{\mathrm{out}}}=0$):
\begin{equation}
\sigma^2_{R_{\mathrm{QPR}}}(q) \;=\; \nu_c(q),
\end{equation}
which is positive whenever correct rollouts differ in process shape; if the judge collapses to a constant value on all correct rollouts of such a group, $\nu_c=0$ and QPR provides no signal on that prompt.

The two mechanisms target the two regimes that outcome-only training cannot resolve: ASR provides signal on routed groups via prefix-length disagreement, recovering gradient on the routed subset with $\sigma^2_{r_{\mathrm{out}}}=0$; QPR provides signal on outcome-uniform-correct groups via heterogeneous trajectory quality, recovering gradient on $\{p_c=1\}$. Both recovered subsets are degenerate under $r_{\mathrm{out}}$.

\section{Protocol Validation: Scaling Discrimination}
\label{sec:appendix_scaling}

Table~\ref{tab:scaling_validation} validates the Step-Quality Evaluation Protocol by applying it to the Qwen3.5 model family \citep{qwen3.5} across five scales. Metrics improve consistently with model capacity, confirming the protocol's discriminative power.

\begin{table*}[htbp]
\centering
\small
\resizebox{\textwidth}{!}{%
\begin{tabular}{l ccccc}
\toprule
\textbf{Metric} & \textbf{Qwen3.5-0.8B} & \textbf{Qwen3.5-2B} & \textbf{Qwen3.5-4B} & \textbf{Qwen3.5-9B} & \textbf{Qwen3.5-27B} \\
\midrule
\multicolumn{6}{l}{\textit{Accuracy (\%) $\uparrow$}} \\
Overall          & 8.7  & 27.0 & 64.4 & 72.5 & 84.2 \\
GPQA\textsubscript{@1}             & 18.7 & 42.9 & 73.2 & 77.3 & 80.3 \\
MATH500\textsubscript{@1}          & 18.1 & 45.9 & 88.7 & 92.5 & 95.5 \\
AIME 2024\textsubscript{@8}        & 1.3  & 17.5 & 62.1 & 71.7 & 86.3 \\
AIME 2025\textsubscript{@8}        & 2.9  & 12.9 & 46.3 & 58.8 & 79.2 \\
\midrule
\multicolumn{6}{l}{\textit{Process Quality}} \\
Useful-step ratio (\%) $\uparrow$       & 50.1 & 61.6 & 64.2 & 63.6 & 63.9 \\
Error-step ratio (\%) $\downarrow$      & 16.5 & 3.5  & 0.7  & 0.6  & 0.3  \\
Redundant-step ratio (\%) $\downarrow$  & 12.3 & 7.6  & 6.6  & 6.9  & 7.6  \\
Reversion-step ratio (\%) $\downarrow$ & 2.1  & 1.8  & 0.9  & 0.6  & 1.0  \\
First-error rel.\ position $\uparrow$  & 0.353 & 0.529 & 0.734 & 0.664 & 0.769 \\
\midrule
\multicolumn{6}{l}{\textit{Efficiency}} \\
Truncation rate (\%) $\downarrow$ & 59.1 & 36.7 & 5.1  & 4.1  & 4.9  \\
Avg.\ steps                       & 27.2 & 28.1 & 29.8 & 29.7 & 31.7 \\
Avg.\ tokens                      & 10{,}606 & 8{,}998 & 5{,}747 & 5{,}238 & 7{,}162 \\
\bottomrule
\end{tabular}
}
\caption{Validation of the Step-Quality Evaluation Protocol across model scales. Metrics are computed on the same evaluation set (GPQA\textsubscript{@1}, MATH500\textsubscript{@1}, AIME 2024\textsubscript{@8}/2025\textsubscript{@8}). Monotonic improvements in accuracy and error-step suppression confirm the protocol's discriminative power.}
\label{tab:scaling_validation}
\end{table*}

\section{Scaffold Decomposition Prompt}
\label{sec:appendix_decompose_prompt}

The system prompt used in the automated decomposition pipeline (Section~\ref{sec:stage1}; statistics in Appendix~\ref{app:scaffold_data_stats}) is shown in Figure~\ref{fig:decompose_prompt}.

\begin{figure}[htbp]
\centering
\small
\begin{tcolorbox}[title=Prompt Template for Scaffold Decomposition, fonttitle=\bfseries\small, boxrule=0.5pt, colback=white, colframe=black]
\small
\textbf{System:} You are a mathematics problem decomposition engine. You will be given a single original math problem and its final correct answer. Your task is to decompose the original problem into several sub-problems.

\medskip
\textbf{Mandatory Rules:}

\textbf{1. Ground-truth validity constraint.} The ground truth of the main problem MUST be a single value, interval, finite set, or algebraic expression. If not, return an empty JSON object \texttt{\{\}}.

\textbf{2. Sub-problem ground truth constraint.} Each sub-problem MUST have a ground truth that is also a single value, interval, finite set, or algebraic expression. Forbidden outputs: ``true/false'', ``holds if and only if ...'', any form of proof or explanation.

\textbf{3. Meaningful decomposition.} Each sub-problem must represent a distinct mathematical concept or technique, require mathematical reasoning (not just arithmetic), and add genuine pedagogical value. Forbidden: pure calculation steps, trivial simplifications.

\textbf{4. Progressive problem chain.} Each sub-problem should use results from previous sub-problems. The last sub-problem must be genuinely different from the main problem. The main problem should require combining insights from multiple sub-problems.

\textbf{5. Problem structure format.}\\
\texttt{[Given conditions and context]}\\
\texttt{Sub-problem 1: [First conceptual step]}\\
\texttt{Sub-problem 2: Using the result of Sub-problem 1, [Second step]}\\
\texttt{...}\\
\texttt{Main Problem: [Final question using results from multiple sub-problems]}

\textbf{6. Answer format specification.} When an answer could be expressed in multiple equivalent ways, append a format requirement: ``Express your answer as [specific format].'' Prefer asking for counts when possible.

\textbf{7. Ground truth LaTeX format constraint.} All answers must be in valid LaTeX math format (without surrounding \$).

\medskip
\textbf{Output JSON structure:}\\
\texttt{\{"data\_source": "step\_math", "prompt": [\{"role": "system", "content": "Solve each sub-problem step by step. Put ONLY the final answer in \textbackslash boxed\{\} labeled [SUB-X ANSWER] / [MAIN ANSWER]."\}, \{"role": "user", "content": "[problem text]"\}], "reward\_model": \{"ground\_truth": \{"sub1": "...", ..., "main": "..."\}, "style": "rule"\}\}}

\medskip
\textbf{Quality check:} (1) Does each sub-problem teach a distinct concept? (2) Is the last sub-problem different from the main problem? (3) Does the main problem require synthesizing multiple results? (4) Would a student learn mathematical thinking? If any answer is ``no'', return \texttt{\{\}}.

\medskip
\textbf{User:} [Question] \quad Answer: [Ground-truth answer]
\end{tcolorbox}
\caption{System prompt for the scaffold decomposition pipeline. The decomposer generates answer-hidden sub-question chains from a given math problem and its ground-truth answer.}
\label{fig:decompose_prompt}
\end{figure}

\section{Concise Reasoning Prompt}
\label{sec:appendix_concise_prompt}

To establish a non-RL baseline for reasoning compression, we prepend the following concise-reasoning instruction to the system prompt when evaluating the base model without any RL training:

\begin{center}
\small
\begin{tcolorbox}[title=Prompt Template for Concise Reasoning, fonttitle=\bfseries\small, boxrule=0.5pt, colback=white, colframe=black]
\small
\textbf{System:} Answer as concisely as possible. Avoid unnecessary redundancy, repetition, and verbose explanations. Focus on key reasoning steps and provide the final answer directly.

\medskip
\textbf{User:} [Question]
\end{tcolorbox}
\captionof{figure}{Concise reasoning prompt prepended at inference time for the non-RL compression baseline.}
\label{fig:concise_prompt}
\end{center}

\section{Step-Quality Evaluation Protocol Details}
\label{app:step_quality_protocol}

\paragraph{Evaluation set instantiation.}
As a concrete instantiation of the protocol, we select four established datasets spanning different reasoning domains and difficulty levels: GPQA \citep{gpqa} for graduate-level scientific reasoning, MATH500 \citep{lightman2023} for competition mathematics, and AIME 2024 \citep{aime24} and AIME 2025 \citep{aime25} for high-difficulty olympiad-style problems. The protocol itself is not tied to these specific datasets and can be applied to any STEM evaluation set with verifiable answers. GPQA and MATH500 are evaluated with single-sample generation. To reduce variance on the smaller and harder AIME subsets, each AIME problem is sampled 8 times from the evaluated model and scores are averaged across rollouts.

\paragraph{Step-level judging protocol.}
For each generated response, final-answer correctness is computed independently using a rule-based answer checker. Process evaluation is applied to the reasoning trace rather than to the final answer, and the LLM Judge is never used to decide whether the final answer is correct. The evaluation judge (Gemini-3-flash-preview \citep{gemini}) is also a different model from the one used to produce QPR training rewards (GPT-4.1-mini), so the protocol's diagnostic measurements are not generated by the same judge that shaped the policy. The judge first parses the reasoning trace into $N$ atomic steps $\{s_1, \dots, s_N\}$ without summarizing or rewriting the model's solution. It then assigns each step a category label $c_i$ from the same five-way taxonomy used in QPR: \emph{useful}, \emph{mechanical}, \emph{redundant}, \emph{reversion}, and \emph{error}.

The five labels separate distinct process behaviors. \emph{Useful} steps make genuine cognitive progress toward the solution. \emph{Mechanical} steps are correct but mostly executional. \emph{Redundant} steps repeat information or re-derive already established facts. \emph{Reversion} steps occur when the model unnecessarily second-guesses or backtracks on previously correct reasoning, causing the solution path to regress. \emph{Error} steps contain locally invalid reasoning, even when the final answer is eventually correct. This taxonomy allows the protocol to distinguish shorter high-quality reasoning from outputs that are merely short, lucky, or under-explained.

\paragraph{Diagnostic metrics.}
The protocol reports decomposed metrics rather than relying only on the aggregate reward used for training. Let $\mathcal{P}$ be the set of evaluated instances, $\hat{a}_p$ the extracted final answer, and $a_p$ the ground-truth answer. We report final-answer accuracy as
\begin{equation}
F = \frac{1}{|\mathcal{P}|}\sum_{p \in \mathcal{P}} \mathbf{1}[\hat{a}_p = a_p],
\end{equation}
with truncated responses counted as incorrect in the overall accuracy setting.

For process quality, we report the useful-step ratio
\begin{equation}
S_u = \frac{1}{N}\sum_i \mathbf{1}[c_i = u],
\end{equation}
and the negative-step ratios
\begin{equation}
r_k = \frac{1}{N}\sum_i \mathbf{1}[c_i = k],
\quad k \in \mathcal{K}.
\end{equation}
Here, $u$ denotes the \emph{useful} label, and $\mathcal{K}=\{\mathrm{mec},\mathrm{red},\mathrm{rev},\mathrm{err}\}$ denotes mechanical, redundant, reversion, and error steps. These metrics directly quantify whether a model uses more substantive reasoning steps and fewer low-value or harmful steps after training.

We additionally measure where the reasoning chain first breaks down. For instances containing at least one error step, we compute
\begin{equation}
S_{\mathrm{first\text{-}err}}
= \frac{1}{|\mathcal{P}_{\mathrm{err}}|}
\sum_{p \in \mathcal{P}_{\mathrm{err}}}
\frac{i^{(p)}_{\mathrm{err}}}{N_p},
\end{equation}
where $i^{(p)}_{\mathrm{err}}$ is the first error-step index and $N_p$ is the number of steps in instance $p$. Higher values indicate that errors occur later in the reasoning chain, reflecting longer locally coherent derivations.

For efficiency, we report average step count and average completion tokens. These efficiency metrics are interpreted jointly with accuracy and useful-step density: a reduction in length is considered beneficial only when final accuracy and useful-step density are preserved or improved.

\paragraph{Evaluation set statistics.}
Table~\ref{tab:app_bench_stats} summarizes the composition of the evaluation set used in this work. In total, it contains 392 unique problems and produces 812 evaluation instances per model.

\begin{table}[htbp]
\centering
\small
\resizebox{\columnwidth}{!}{%
\begin{tabular}{l r r r}
\toprule
\textbf{Dataset} & \textbf{Problems} & \textbf{Samples} & \textbf{Total} \\
\midrule
GPQA        & 198 & 1 & 198 \\
MATH500     & 134 & 1 & 134 \\
AIME 2024   & 30  & 8 & 240 \\
AIME 2025   & 30  & 8 & 240 \\
\midrule
\textbf{Total} & 392 & --- & 812 \\
\bottomrule
\end{tabular}
}
\caption{Composition of the evaluation set. AIME problems are sampled multiple times to reduce evaluation variance on high-difficulty olympiad problems.}
\label{tab:app_bench_stats}
\end{table}

\paragraph{Implementation pipeline.}
The protocol is implemented as a three-stage pipeline. We deliberately use a different model for evaluation than for QPR training: all evaluation-time judging (step parsing, step scoring, and the IAA study in Appendix~\ref{app:human_agreement}) uses \textbf{Gemini-3-flash-preview}, whereas QPR training rewards are produced by GPT-4.1-mini (Appendix~\ref{app:qpr_reward_details}). This decoupling ensures that the diagnostic metrics reported in the main results are not produced by the same model that shaped the policy during training. Given a set of problems with ground-truth answers and a target model to evaluate, the pipeline proceeds as follows:
\begin{enumerate}
\item \textbf{Model Rollout.} Each problem is sent to the target model. The model generates a complete reasoning trace followed by a final answer. Responses that exceed the maximum token budget are marked as truncated and counted as incorrect in overall accuracy, but excluded from process-quality evaluation (since their reasoning chains are incomplete).

\item \textbf{Step Parsing.} An LLM Judge (Gemini-3-flash-preview) segments each non-truncated reasoning trace into atomic steps. The judge is instructed to faithfully reconstruct the original reasoning process without summarizing, merging, or optimizing --- preserving redundancy, hesitation, and self-correction as separate steps.

\item \textbf{Step Scoring.} A second LLM Judge call evaluates each parsed step by assigning a mutually exclusive category label applied in strict priority order (reversion $>$ error $>$ redundant $>$ mechanical $>$ useful). The judge also renders a binary final-answer correctness decision by comparing the model's answer against the ground truth.
\end{enumerate}
All metrics reported in the main paper are computed from the Step Scoring outputs: category ratios are step-count fractions, and first-error position is computed per-problem then averaged.

\subsection{Step Parsing Prompt}

The following prompt instructs the LLM Judge to segment a model's reasoning trace into atomic steps without any modification or judgment.

\begin{center}
\small
\begin{tcolorbox}[title=Prompt Template for Step Parsing, fonttitle=\bfseries\small, boxrule=0.5pt, colback=white, colframe=black]
\small
\textbf{System:} You are a mathematical reasoning analyst.

Your goal is to faithfully reconstruct the ORIGINAL reasoning process step by step, NOT to summarize or clean it.

\medskip
\textbf{Critical Rules:}
\begin{itemize}\setlength{\itemsep}{1pt}
\item Preserve redundancy: if the model repeats the same idea, create multiple steps.
\item Preserve verification: include sanity checks, examples, and side explorations.
\item Preserve hesitation and re-explanations.
\item DO NOT merge similar steps.
\item DO NOT optimize or simplify the reasoning.
\item DO NOT judge whether any step is correct, useful, or redundant --- just segment faithfully.
\end{itemize}

Each step should correspond to ONE actual reasoning action in the original text.

\medskip
\textbf{Output format:}\\
\texttt{\{"steps": [\{"step\_id": 0, "content": "what the model is doing in this exact moment"\}, ...]\}}

\medskip
\textbf{User:} Problem: [Question]

Model output: [Full model reasoning trace]
\end{tcolorbox}
\captionof{figure}{Prompt template for the step parsing stage. The judge segments each reasoning trace into atomic steps without summarization or modification.}
\label{fig:step_parsing_prompt}
\end{center}

\subsection{Step Scoring Prompt}

The following prompt instructs the LLM Judge to assign category labels to each parsed step, and to determine final-answer correctness.

\begin{center}
\small
\begin{tcolorbox}[title=Prompt Template for Step Scoring, fonttitle=\bfseries\small, boxrule=0.5pt, colback=white, colframe=black]
\small
\textbf{System:} You are an expert evaluator of mathematical reasoning quality. You will be given: (1) a math problem, (2) the ground truth answer, (3) the model's full reasoning output and its structured reasoning steps.

Evaluate on three dimensions:

\medskip
\textbf{Dimension 1 --- Per-step Category} (mutually exclusive, applied in priority order):

\begin{enumerate}\setlength{\itemsep}{1pt}
\item \textbf{reversion} --- The model was on a correct reasoning path but unnecessarily second-guesses, backtracks, or revises a previously correct step, causing the reasoning to regress (look for ``wait'', ``actually'', ``let me reconsider'', ``I think I was wrong'' when the original reasoning was in fact correct).
\item \textbf{error} --- The step contains a mathematical or logical mistake NOT accompanied by explicit self-correction (silently wrong).
\item \textbf{redundant} --- Mathematically correct but merely restates something already established, adding no new information.
\item \textbf{mechanical} --- Correct and non-redundant, but purely executional with no new cognitive contribution (e.g., expanding algebra, substituting values, arithmetic).
\item \textbf{useful} --- Everything else: genuine cognitive contribution that advances the solution (choosing strategy, identifying insight, applying non-trivial theorem, case-splitting).
\end{enumerate}

\medskip
\textbf{Dimension 2 --- Answer Correctness:}\\
Compare model's final answer with ground truth. Set \texttt{is\_correct} to true/false.

\medskip
\textbf{Output format:}\\
\texttt{\{"is\_correct": true, "per\_step\_scores": [\{"step\_id": 0, "category": "useful"\}, ...]\}}

\medskip
\textbf{User:} Problem: [Question]

Ground truth answer: [Answer]

Model's full output: [Reasoning trace]

Model's structured reasoning steps: [Parsed steps from Stage 2]
\end{tcolorbox}

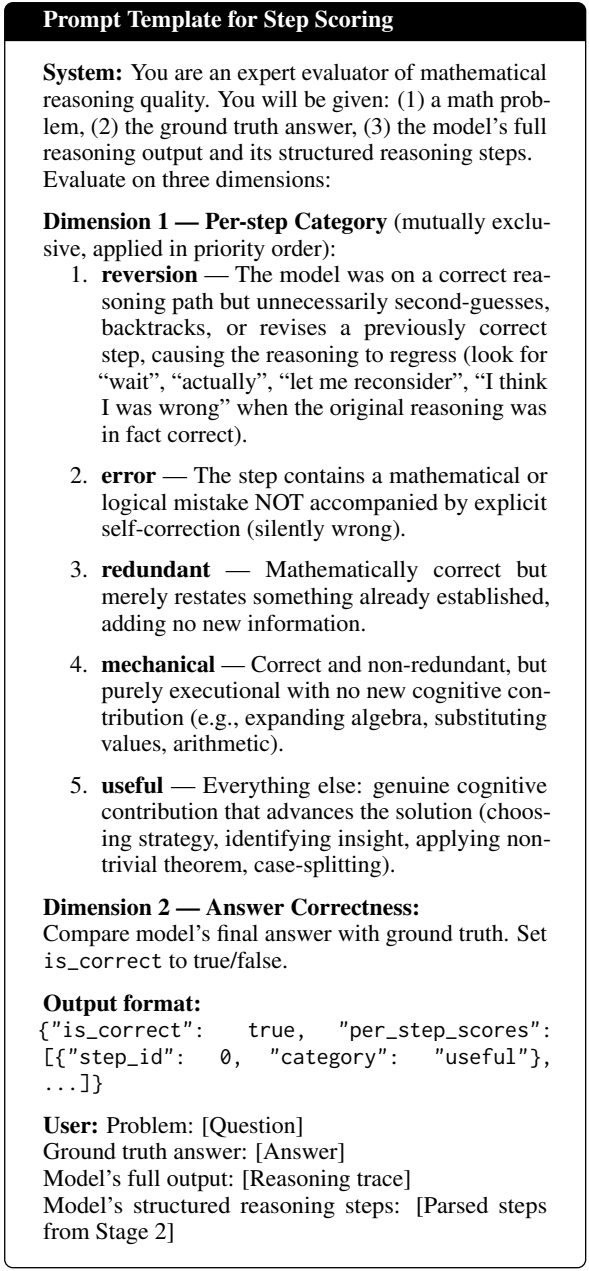
\captionof{figure}{Prompt template for the step scoring stage. The judge assigns a category label to each parsed step and determines final-answer correctness.}
\label{fig:step_scoring_prompt}
\end{center}

\section{QPR Reward Details}
\label{app:qpr_reward_details}

\paragraph{Step labels.}
QPR uses five mutually exclusive labels to characterize the role of each reasoning step after final-answer correctness has been verified. \emph{Useful} steps make genuine cognitive progress toward the solution. \emph{Mechanical} steps are correct but primarily executional computations. \emph{Redundant} steps restate or re-derive information that has already been established. \emph{Reversion} steps occur when the model unnecessarily second-guesses or backtracks on previously correct reasoning, causing the solution path to regress. \emph{Error} steps contain locally invalid reasoning even when the final answer is eventually correct.

\paragraph{Judge annotation.}
For each correct-answer rollout, the LLM Judge parses the trace into atomic reasoning steps without rewriting the solution and assigns one category label to each step. These annotations are used only to construct the process-quality reward; final-answer correctness remains determined by the rule-based verifier.

\paragraph{Judge prompt template.}
The LLM Judge (GPT-4.1-mini) receives each problem, its ground-truth answer, and the model's complete reasoning trace; it returns a structured JSON with per-step category labels.

\paragraph{Reward hyperparameters.}
In our experiments, we set $\alpha=0.5$. The penalty weights are $\lambda_{\mathrm{mec}}=0.05$, $\lambda_{\mathrm{red}}=0.20$, $\lambda_{\mathrm{rev}}=0.25$, and $\lambda_{\mathrm{err}}=0.40$, assigning the strongest penalty to locally invalid reasoning and smaller penalties to low-value but correct computation.

\section{Training Implementation Details}
\label{app:training_details}

All experiments are conducted on a single node equipped with 8 NVIDIA H800 GPUs, using the verl~0.7.0 training framework with vLLM~0.9.2 as the rollout engine. The base model is \textbf{Qwen3-8B-Instruct} \citep{qwen3}. Both stages use GRPO \citep{grpo} with $n{=}8$ rollout samples per prompt, a prompt batch size of 48, and a maximum prompt length of 1,024 tokens. KL penalty is disabled in both stages, including both reward-side KL and loss-side KL loss, and the entropy coefficient is set to 0.

\paragraph{Stage 1 --- Adaptive Scaffolded RL.}
Each batch begins with rollouts on the original problems under the  outcome reward. Any group whose mean outcome reward falls below the threshold $\tau$ is re-rolled out on the cached scaffolded prompt and scored with the prefix-consistent scaffold reward defined in Section~\ref{sec:stage1}. Within a routed group, scaffolded rollouts \emph{replace} rather than augment the original rollouts, so each parameter update aggregates the same number of loss-contributing trajectories ($G{=}8$ per prompt) as outcome-only GRPO; this preserves compute and gradient-batch parity with the baseline. Scaffold chains are generated offline with a teacher decomposer and cached on disk; no decomposition model is invoked during RL. The retained original-problem trajectories and the scaffolded trajectories are combined into a single batch for the GRPO \citep{grpo} parameter update. No separate value network is used; the mixed reward serves directly as the scalar signal for advantage estimation. The learning rate is set to $1 \times 10^{-6}$ with a 60-step linear warmup and no decay, and the maximum response length is 8{,}192 tokens.

\paragraph{Stage 2 --- Quality-Aware Process RL.}
QPR is initialized from the best ASR checkpoint and trained with a lower learning rate of $5 \times 10^{-7}$ under a cosine decay schedule. The maximum response length is extended to 16,384 tokens to accommodate longer reasoning traces during the early phase of QPR training. Each rollout is scored by the correctness-gated process reward defined in Eq.~\eqref{eq:qpr_process}: incorrect final answers receive zero reward, while correct-answer rollouts are annotated by an LLM Judge implemented using GPT-4.1-mini. The judge decomposes each reasoning trace into atomic steps and assigns one of five mutually exclusive labels (\emph{useful, mechanical, redundant, reversion, error}).

\section{Scaffold Routing Threshold Sensitivity}
\label{app:routing_threshold}

Table~\ref{tab:routing_threshold} reports the effect of the scaffold routing threshold $\tau$ on Stage~1 (ASR) performance. We vary $\tau \in \{0.25, 0.5, 0.75\}$ across both training sources and report per-benchmark accuracy and average accuracy. The default threshold used in the main experiments is $\tau=0.5$.

\begin{table*}[htbp]
\centering
\small
{\setlength{\tabcolsep}{12pt}%
\resizebox{\textwidth}{!}{%
\begin{tabular}{@{\hskip 1.5em} l c ccccc @{\hskip 1.5em}}
\toprule
\textbf{Training Data} & $\boldsymbol{\tau}$ & \textbf{GPQA\textsubscript{@1}} & \textbf{MATH500\textsubscript{@1}} & \textbf{AIME24\textsubscript{@8}} & \textbf{AIME25\textsubscript{@8}} & \textbf{Avg} \\
\midrule
\multirow{3}{*}{DAPO-Math}
& 0.25 & 51.52 & 89.55 & 67.08 & 47.08 & 63.81 \\
& 0.50 & 51.52 & \textbf{90.23} & 65.00 & \textbf{56.67} & \textbf{65.86} \\
& 0.75 & 51.52 & 85.71 & 63.75 & 55.42 & 64.10 \\
\midrule
\multirow{3}{*}{Big-Math}
& 0.25 & 40.91 & 87.97 & 57.92 & 50.00 & 59.20 \\
& 0.50 & \textbf{53.03} & 87.22 & \textbf{59.17} & \textbf{50.00} & \textbf{62.36} \\
& 0.75 & 50.00 & 85.71 & 45.42 & 35.83 & 54.24 \\
\bottomrule
\end{tabular}
}
}
\caption{Sensitivity of ASR (Stage~1) to the scaffold routing threshold $\tau$. Bold indicates the best average accuracy within each training-data group. $\tau=0.5$ achieves the best average accuracy on both data sources.}
\label{tab:routing_threshold}
\end{table*}

\section{ASR Reward-Signal Ablation}
\label{app:asr_reward_ablation}

To separate scaffold-provided structure from reward densification, Table~\ref{tab:asr_reward_ablation} compares two additional ASR variants on the same scaffolded prompts. \textit{Scaffold + Final Reward} keeps the scaffolded input format but rewards only the final main answer, removing all sub-answer rewards; it improves only modestly over outcome-only GRPO and remains far below ASR, indicating that the scaffold prompt itself is not the main source of the gain. \textit{Scaffold + Independent Rewards} rewards each correct sub-answer independently, without the prefix-consistency gate; it performs better than final-only scaffolding but remains below ASR, showing that dense sub-answer rewards are important and that enforcing prerequisite order further improves credit assignment.

\begin{table}[t]
\centering
\scriptsize
\setlength{\tabcolsep}{3pt}
\resizebox{\columnwidth}{!}{
\begin{tabular}{lccc}
\toprule
\textbf{Variant} & \textbf{Scaffold} & \textbf{Sub-answer reward} & \textbf{Avg.} \\
\midrule
GRPO & -- & -- & 55.80 \\
Scaffold + Final Reward & \checkmark & -- & 57.63 \\
Scaffold + Independent Rewards & \checkmark & independent & 60.19 \\
\textbf{ASR only} & \checkmark & prefix-consistent & \textbf{65.86} \\
\bottomrule
\end{tabular}
}
\caption{Reward-signal ablation for ASR. Scaffold + Final Reward uses the same answer-hidden scaffolded prompts as ASR but keeps only the final main-answer reward. Scaffold + Independent Rewards rewards each scaffolded sub-answer independently, removing the prefix-consistency gate. Avg. is the arithmetic mean of GPQA, MATH500, AIME24, and AIME25 accuracies under the same evaluation protocol as Table~\ref{tab:ablation_stage1}.}
\label{tab:asr_reward_ablation}
\end{table}

\section{Scaffold Construction Statistics}
\label{app:scaffold_data_stats}

Table~\ref{tab:app_data_stats} summarizes the coverage and granularity of the automated scaffold construction pipeline across the two training sources. ``Filtered'' refers to problems automatically excluded because their ground-truth answers do not satisfy the verifiability constraint, such as proof-based or open-ended problems. ``Avg.\ Subs'' denotes the average number of sub-problems per successfully decomposed question.

\begin{table*}[htbp]
\centering
\small
\resizebox{\linewidth}{!}{
\begin{tabular}{l r r r r r r r r}
\toprule
\textbf{Dataset} & \textbf{Total} & \textbf{Success} & \textbf{Filtered} & \textbf{Failed} & \textbf{Succ.\ Rate} & \textbf{Avg.\ Subs} & \textbf{Max} & \textbf{Min} \\
\midrule
DAPO-Math (2.4K) & 2,400 & 2,382 & 16  & 2 & 99.3\% & 3.84 & 13 & 2 \\
DAPO-Math (full) & 17,406 & 17287 & 116  & 3 & 99.3\% & 3.80 & 15 & 1 \\
Big-Math (2.4K)  & 2,400 & 2,251 & 149 & 0 & 93.8\% & 3.01 & 9  & 1 \\
Big-Math (full)  & 12,400 & 11,846 & 554 & 0 & 95.5\% & 2.97 & 11 & 1 \\
\bottomrule
\end{tabular}
}
\caption{Statistics of the automated decomposition pipeline across datasets.}
\label{tab:app_data_stats}
\end{table*}

\section{Human Validation of Scaffold Prerequisite Chains}
\label{app:scaffold_human_validation}

To verify that the generated scaffolds genuinely form prerequisite chains rather than loosely related sub-questions, we randomly sampled 200 scaffolds from the training data and conducted manual annotation. Two annotators independently examined each scaffold and judged whether every later sub-question explicitly depends on the result of at least one earlier sub-question (i.e., no dependency breakage or redundant sub-questions that bypass the chain). A scaffold is marked invalid if any posterior sub-question can be answered without using the result of its designated predecessor.

All 200 inspected scaffolds (100\%) satisfy strict prerequisite-chain integrity: each sub-question builds upon previous sub-question results, and no dependency breakage is observed. This confirms that the automated validity filter (Eq.~\eqref{eq:scaffold_valid}) effectively ensures well-formed prerequisite structures before RL training.

\section{Protocol Validation: Step-Label Agreement with Human Experts}
\label{app:human_agreement}

The Step-Quality Evaluation Protocol's headline metric (\emph{useful}-step ratio $S_u$) and QPR's positive reward term both depend on the binary distinction between \emph{useful} steps and the four low-value categories. We therefore directly validate this binary distinction against human experts.

\paragraph{Annotation setup.}
We randomly sample 50 reasoning trajectories generated by Qwen3-8B-Instruct (the model used in our main experiments) across our evaluation benchmarks (GPQA, MATH500, AIME). All trajectories are pre-parsed into atomic steps by the same LLM Judge used in evaluation; human annotators see the trajectory with step boundaries already drawn and do not re-segment. We deliberately scope this study to label-level agreement \emph{conditional on} the protocol's segmentation, since this matches how the protocol is used at evaluation time; segmentation reliability is checked separately by the scaling-discrimination study (Appendix~\ref{sec:appendix_scaling}), where the protocol cleanly separates five model sizes. The 50 trajectories contain approximately 1{,}250 atomic steps in total (mean 25 steps per trajectory), approximately balanced across the three benchmarks.

\paragraph{Asymmetric annotation protocol.}
To control annotation cost while keeping the LLM Judge consistent with how it is actually used during evaluation, we adopt an asymmetric setup:
\begin{itemize}
    \item \textbf{Human annotators} mark each step as \emph{useful} or \emph{non-useful} only. Two domain experts with PhD-level training in mathematics and physics annotate independently. We do not require them to discriminate among the four low-value categories, since the binary distinction (a)~determines the protocol's headline metric $S_u$ and the sign of QPR's positive reward, (b)~is the boundary that experts can adjudicate quickly and reliably, and (c)~is the only process-quality distinction we \emph{report and analyze} as an evaluation metric. The four low-value categories do shape the QPR \emph{training} reward through their penalty weights (Eq.~\eqref{eq:qpr_penalty}); we therefore do not claim their boundaries are irrelevant, only that validating the reported diagnostic ($S_u$) requires validating the useful/non-useful boundary, which is what this study targets.
    \item \textbf{LLM Judge} produces the same five-class labels (\emph{useful}, \emph{mechanical}, \emph{redundant}, \emph{reversion}, \emph{error}) it produces during evaluation, with no special configuration for this study. For comparison against the human binary labels, we collapse the four low-value classes into \emph{non-useful}; this is the same aggregation convention used in QPR's $\Phi$ penalty and in reporting the useful-step ratio $S_u$.
\end{itemize}
Annotators receive the same rubric used by the LLM Judge (Appendix~\ref{app:step_quality_protocol}) but are blinded to the LLM Judge's outputs and to which method produced each trajectory.

\paragraph{Step-level binary agreement.}
Table~\ref{tab:human_iaa_overall} reports binary agreement between expert consensus and the LLM Judge, with inter-annotator agreement (IAA) between the two human experts as the human ceiling. Expert consensus is defined as the agreed label when both experts match (covering 90.0\% of steps); on the remaining 10.0\% the LLM is compared against each expert separately and scored values are averaged.

\begin{table*}[h]
\centering\small
{\setlength{\tabcolsep}{35pt}%
\resizebox{\textwidth}{!}{%
\begin{tabular}{@{\hskip 1.5em} lcc @{\hskip 1.5em}}
\toprule
\textbf{Pair} & \textbf{Accuracy} & \textbf{Cohen's $\kappa$} \\
\midrule
Expert A vs.\ Expert B (IAA, ceiling) & 0.90 & 0.79 \\
LLM Judge vs.\ Expert consensus       & 0.88 & 0.75 \\
LLM Judge vs.\ Expert A               & 0.87 & 0.73 \\
LLM Judge vs.\ Expert B               & 0.88 & 0.75 \\
\bottomrule
\end{tabular}
}
}
\caption{Step-level useful/non-useful binary agreement on 1{,}250 steps from 50 Qwen3-8B-Instruct trajectories. The LLM Judge here refers to the evaluation judge (Gemini-3-flash-preview \citep{gemini}), the same model used to produce all diagnostic metrics in the main results. Cohen's $\kappa$ is within $0.04$ of the human ceiling, indicating that the protocol's core useful/non-useful distinction is approximately as consistent with each expert as the experts are with each other.}
\label{tab:human_iaa_overall}
\end{table*}

\paragraph{Per-class binary metrics.}
Table~\ref{tab:human_iaa_per_class} reports per-class precision/recall/F1 of the LLM Judge against expert consensus for the \emph{useful} and \emph{non-useful} classes, alongside the corresponding inter-annotator F1.

\begin{table}[h]
\centering\small
\resizebox{\columnwidth}{!}{%
\begin{tabular}{lcccc}
\toprule
\textbf{Class} & \textbf{Precision} & \textbf{Recall} & \textbf{F1} & \textbf{IAA F1} \\
\midrule
useful      & 0.88 & 0.90 & 0.89 & 0.91 \\
non-useful  & 0.87 & 0.85 & 0.86 & 0.89 \\
\bottomrule
\end{tabular}
}
\caption{Per-class agreement of the LLM Judge against expert consensus on 1{,}250 steps. The IAA F1 column reports the corresponding per-class F1 between the two human experts. The LLM Judge's F1 is within $0.02$--$0.03$ of the human ceiling on both classes, supporting the use of the useful-step ratio $S_u$ as a reliable diagnostic metric and as input to QPR's positive reward term.}
\label{tab:human_iaa_per_class}
\end{table}

\paragraph{Scope of this validation.}
This study validates the \emph{protocol} only: it shows that the LLM Judge's per-step useful/non-useful labels are consistent with expert labels at near-ceiling levels. The pairwise preference results in Section~\ref{sec:analysis_and_cases} are a separate piece of evidence about \emph{model output quality} (SCOPE-RL vs.\ GRPO under expert preference) and are not used here as protocol-validation evidence.

\section{Pairwise Expert Evaluation}
\label{app:llm_judge_eval}

Table~\ref{tab:app_llm_judge} reports the full dimension-level results of the pairwise expert evaluation. Three domain experts (Ph.D.-level mathematicians) independently judge 200 examples where both GRPO and SCOPE-RL produce the correct final answer. For each example and dimension, the majority vote among the three annotators determines the winner.

\begin{table}[htbp]
\centering
\small
\resizebox{\columnwidth}{!}{%
\begin{tabular}{lcc}
\toprule
\textbf{Dimension} & \textbf{SCOPE-RL} & \textbf{GRPO} \\
\midrule
Clarity            & 131 & 69 \\
Conciseness        & 144 & 56 \\
Non-redundancy     & 138 & 62 \\
Logical Coherence  & 119 & 81 \\
Overall            & 141 & 59 \\
\bottomrule
\end{tabular}
}
\caption{Pairwise expert evaluation (count out of 200) on examples where both GRPO and SCOPE-RL produce the correct final answer. Winner is determined by majority vote among three domain experts.}
\label{tab:app_llm_judge}
\end{table}

\section{Robustness Results: GSPO Backend and 0.6B Scale}
\label{app:robustness_results}

Table~\ref{tab:robustness} reports results when SCOPE-RL is instantiated with the GSPO optimizer instead of GRPO (panel a), and on the smaller Qwen3-0.6B-Instruct model (panel b). Both experiments use DAPO-Math as the training source and follow the same two-stage protocol as the main experiments.
The GSPO results (Table~\ref{tab:robustness}a) confirm that SCOPE-RL's gains are not tied to a single optimizer implementation: ASR improves accuracy from 61.60\% to 66.04\%, and QPR further raises it to 66.93\% while improving the useful-step ratio from 60.89\% to 63.27\% and reducing token usage by 13.1\%, both relative to the GSPO baseline. The 0.6B results (Table~\ref{tab:robustness}b) show that even under limited model capacity, the stage-wise decomposition remains effective: ASR lifts average accuracy from 26.06\% to 30.39\%, and QPR improves it to 32.06\% while raising Useful\% from 51.79\% to 58.43\% and reducing tokens by 22.9\%, both relative to the GRPO baseline.

\onecolumn
\begin{table*}[t!]
\centering
\small
\setlength{\tabcolsep}{4pt}
\resizebox{\textwidth}{!}{%
\begin{tabular}{l ccccc cc r}
\toprule
\multirow{2}{*}{\textbf{Method}}
  & \multicolumn{5}{c}{\textbf{Accuracy (\%) $\uparrow$}}
  & \multicolumn{2}{c}{\textbf{Reasoning Quality $\uparrow$}}
  & \multirow{2}{*}{\textbf{Avg.\ Tokens $\downarrow$}} \\
\cmidrule(lr){2-6}\cmidrule(lr){7-8}
& GPQA\textsubscript{@1} & MATH500\textsubscript{@1} & AIME24\textsubscript{@8} & AIME25\textsubscript{@8} & Avg & Useful\% & 1st-Err-Pos & \\
\midrule
\multicolumn{9}{l}{\textit{(a) GSPO optimizer on Qwen3-8B-Instruct}} \\
\midrule
GSPO               & 48.48 & 85.82 & 57.08 & 55.00 & 61.60 & 60.89 & 0.736 & 6{,}719 \\
\addlinespace[3pt]
ASR only            & \textbf{50.51} & 90.30 & 65.83 & 57.50 & 66.04 & 58.51 & 0.697 & 7{,}460 \\
SCOPE-RL            & 50.00 & \textbf{91.04} & \textbf{67.92} & \textbf{58.75} & \textbf{66.93} & \textbf{63.27} & \textbf{0.752} & \textbf{5{,}836} \\
\midrule
\multicolumn{9}{l}{\textit{(b) GRPO optimizer on Qwen3-0.6B-Instruct}} \\
\midrule
Qwen3-0.6B-Instruct & 20.20 & 32.84 & 3.75 & 8.75 & 16.39 & 40.24 & 0.393 & 3{,}467 \\
\midrule
GRPO               & 15.66 & 51.49 & 14.58 & 22.50 & 26.06 & 51.79 & 0.574 & 10{,}188 \\
\addlinespace[3pt]
ASR only            & 20.20 & 59.70 & 16.67 & 25.00 & 30.39 & 55.15 & 0.579 & 9{,}367 \\
SCOPE-RL            & \textbf{21.72} & \textbf{61.94} & \textbf{18.33} & \textbf{26.25} & \textbf{32.06} & \textbf{58.43} & \textbf{0.624} & \textbf{7{,}852} \\
\bottomrule
\end{tabular}
}
\caption{Robustness checks on alternative optimizer and model scale (both trained with DAPO-Math). (a)~GSPO optimizer on Qwen3-8B-Instruct: SCOPE-RL's two-stage gains transfer to an alternative RL optimizer. (b)~GRPO optimizer on Qwen3-0.6B-Instruct: despite limited model capacity, the two-stage decomposition yields consistent gains in accuracy, reasoning quality, and efficiency.}
\label{tab:robustness}
\end{table*}

\section{Case Study: Reasoning Trace Comparison}
\label{sec:appendix_case_study}

\definecolor{cU}{rgb}{0.10, 0.50, 0.22}
\definecolor{cE}{rgb}{0.81, 0.13, 0.18}
\definecolor{cR}{rgb}{0.75, 0.53, 0.00}
\definecolor{cF}{rgb}{0.51, 0.31, 0.87}
\definecolor{cM}{rgb}{0.43, 0.47, 0.51}

\begin{tcolorbox}[
  colback=gray!5, colframe=black!40, boxrule=0.3pt,
  left=3pt, right=3pt, top=2pt, bottom=2pt]
\footnotesize
\textbf{Problem.}\;
Let $a,b,c$ be distinct real numbers. Find the degree of\;
$p(x)=\tfrac{(x-b)(x-c)}{(a-b)(a-c)}+\tfrac{(x-a)(x-c)}{(b-a)(b-c)}+\tfrac{(x-a)(x-b)}{(c-a)(c-b)}.$
\;\textbf{Answer:}\;$0$.
\end{tcolorbox}

\vspace{3pt}

\noindent
\begin{minipage}[t]{0.34\textwidth}
\vspace{0pt}%
\begin{tcolorbox}[
  title={\centering\footnotesize\textbf{GRPO (Baseline)}},
  colback=white, colframe=black!60, boxrule=0.4pt,
  fonttitle=\bfseries, coltitle=white, colbacktitle=black!70,
  top=2pt, bottom=2pt, left=2pt, right=2pt]
\tiny

\textcolor{cM}{\ldots\;We are to find the \textbf{degree} of the polynomial $p(x)$, given that $a,b,c$ are distinct real numbers. Let me analyze this expression.\;\ldots}

\smallskip
\textcolor{cU}{This expression resembles the \textbf{Lagrange interpolation polynomial}.}

\smallskip
\textcolor{cE}{So if we take $f(x)=x^2$, then this is the Lagrange interpolation of $x^2$ at three points.\;Since interpolating a degree-2 polynomial with three points gives exactly that polynomial:\;$p(x)=x^2$.\;Therefore, the degree is \textbf{2}.}

\medskip
\textcolor{cM}{But let me verify this with an example.\;Let me take $a{=}0,\,b{=}1,\,c{=}{-}1$.}

\smallskip
\textcolor{cM}{First term:\;$\tfrac{(x-1)(x+1)}{(-1)(1)}=-x^2+1$;\;Second:\;$\tfrac{x(x+1)}{2}=\tfrac{x^2+x}{2}$;\;Third:\;$\tfrac{x(x-1)}{2}=\tfrac{x^2-x}{2}$}

\smallskip
\textcolor{cM}{Add:\;$(-x^2+1)+\tfrac{2x^2}{2}=-x^2+1+x^2=1$}

\medskip
\textcolor{cF}{Wait, that gives 1?\;But that contradicts degree 2.\;So something is wrong with my reasoning.}

\smallskip
\textcolor{cF}{Why?\;If $f(x)=x^2$ and we interpolate with 3 points, it should give $x^2$\ldots\;but the result is 1?\;Wait, maybe I confused $f(x)$?}

\medskip
\textcolor{cR}{Let me check $p(a),p(b),p(c)$.\;\ldots\;All equal 1.\;So $p(x)$ is 1 at three distinct points.}

\smallskip
\textcolor{cF}{But a polynomial of degree $\le 2$ that equals 1 at three distinct points must be the constant 1.\;But that would mean $p(x)=1$, degree 0.\;That contradicts my earlier conclusion\ldots\;Let me reconsider.}

\smallskip
\textcolor{cF}{But why did I think it was $x^2$?}

\medskip
\textcolor{cU}{I confused the Lagrange interpolation of $f(x)=x^2$ with the sum of basis polynomials.\;In our case, $p(x)=L_a(x)+L_b(x)+L_c(x)=1$.\;Because for any $x$, the sum of the basis polynomials is 1.\;So $p(x)=1$, degree is 0.}

\medskip
\textcolor{cR}{But why is the problem non-trivial?\;Maybe the problem is different?\;\ldots\;Assuming correct, the answer is 0.\;\ldots\;Let me check once more.\;\ldots}

\medskip
\textcolor{cU}{So the answer is 0.\;Therefore, the degree is $\boxed{0}$.}

\end{tcolorbox}
\end{minipage}%
\hfill%
\begin{minipage}[t]{0.32\textwidth}
\vspace{0pt}%
\begin{tcolorbox}[
  title={\centering\footnotesize\textbf{Stage~1 (ASR)}},
  colback=white, colframe=black!60, boxrule=0.4pt,
  fonttitle=\bfseries, coltitle=white, colbacktitle=teal!70!black,
  top=2pt, bottom=2pt, left=2pt, right=2pt]
\tiny

\textcolor{cM}{\ldots\;We are to find the degree of this polynomial, given that $a,b,c$ are distinct real numbers.\;\ldots}

\smallskip
\textcolor{cR}{Each numerator is a quadratic in $x$; each denominator is a constant (since $a,b,c$ are constants).\;So each term is a quadratic polynomial.\;But when added together, the sum may simplify to a lower degree.}

\smallskip
\textcolor{cR}{The question is: what is the degree of the resulting polynomial?}

\medskip
\textcolor{cU}{This expression resembles the \textbf{Lagrange interpolation polynomial}.\;Specifically, it is the sum of the Lagrange basis polynomials for interpolation at three points $(a,1),(b,1),(c,1)$.}

\smallskip
\textcolor{cU}{A quadratic polynomial taking the value 1 at three distinct points must be the constant 1.\;Therefore, $p(x)=1$, and the degree is 0.}

\medskip
\textcolor{cF}{But wait, that seems surprising.\;Let me verify with an example.}

\smallskip
\textcolor{cM}{$a{=}0,\,b{=}1,\,c{=}2$.\;First:\;$\tfrac{(x-1)(x-2)}{2}$;\;Second:\;$-x^2+2x$;\;Third:\;$\tfrac{x(x-1)}{2}$}

\smallskip
\textcolor{cM}{Combine:\;$\tfrac{(x^2-3x+2)+(-2x^2+4x)+(x^2-x)}{2}=\tfrac{2}{2}=1$}

\medskip
\textcolor{cU}{So indeed, $p(x)=1$, a constant polynomial.\;Therefore, the degree is 0.}

\smallskip
\textcolor{cR}{So the degree of the polynomial is $\boxed{0}$.}

\end{tcolorbox}
\end{minipage}%
\hfill%
\begin{minipage}[t]{0.32\textwidth}
\vspace{0pt}%
\begin{tcolorbox}[
  title={\centering\footnotesize\textbf{Stage~2 (QPR)}},
  colback=white, colframe=black!60, boxrule=0.4pt,
  fonttitle=\bfseries, coltitle=white, colbacktitle=blue!60!black,
  top=2pt, bottom=2pt, left=2pt, right=2pt]
\tiny

\textcolor{cM}{\ldots\;We are to find the degree of the polynomial $p(x)$.\;\ldots}

\smallskip
The denominators are constants (since $a,b,c$ are fixed), and the numerators are quadratic polynomials in $x$.\;So each term is a quadratic polynomial divided by a constant.\;\ldots\;when we add them together, the result may simplify to a lower-degree polynomial.

\medskip
\textcolor{cU}{Specifically, if we have three distinct points $a,b,c$, then the \textbf{Lagrange basis polynomials} are:}
\begin{gather*}
{\scriptstyle L_a(x)=\tfrac{(x-b)(x-c)}{(a-b)(a-c)},}\\[-2pt]
{\scriptstyle L_b(x)=\ldots,\;L_c(x)=\ldots}
\end{gather*}

\textcolor{cU}{Then the sum $L_a(x)+L_b(x)+L_c(x)$ is the sum of the Lagrange basis polynomials for interpolation at three points.\;But the sum of all Lagrange basis polynomials is the constant polynomial 1, because for any $x$, the sum of $L_a(x)+L_b(x)+L_c(x)$ equals 1.}

\medskip
\textcolor{cU}{Therefore, the given expression $p(x)$ is equal to 1 for all $x$, hence it is the constant polynomial 1.\;Therefore, the degree is 0.}

\medskip
\textcolor{cM}{Quick check:\;$a{=}0,b{=}1,c{=}2$.\;\ldots\;numerator is 2, denominator is 2:\;$p(x)=1$.\;\checkmark}

\medskip
\textcolor{cU}{So indeed, $p(x)=1$, a constant polynomial.\;Therefore, the degree of $p(x)$ is $\boxed{0}$.}

\end{tcolorbox}
\end{minipage}

\vspace{5pt}

\centering\footnotesize
\colorbox{cU!15}{\textcolor{cU}{\textbf{Useful insight}}}\;\;
\colorbox{cE!15}{\textcolor{cE}{\textbf{Error}}}\;\;
\colorbox{cF!15}{\textcolor{cF}{\textbf{Reversion}}}\;\;
\colorbox{cR!15}{\textcolor{cR}{\textbf{Redundant}}}\;\;
\colorbox{cM!15}{\textcolor{cM}{\textbf{Mechanical / omitted}}}

\end{document}